\documentclass[10pt,twocolumn,letterpaper]{article}

\usepackage{iccv}              % To produce the CAMERA-READY version
\usepackage{amsmath}
\usepackage{mathtools}
\usepackage{amsthm}
\usepackage{algorithm}
\usepackage[noend]{algpseudocode}
\usepackage{graphicx}
\usepackage{blindtext}
\usepackage{caption}
\usepackage{svg}
\usepackage{longtable}
\usepackage{colortbl}
\usepackage{soul}
\usepackage{pifont}
\usepackage{stackengine}
\usepackage[accsupp]{axessibility}
\usepackage{xr}
\newtheorem{theorem}{Theorem} % 定理编号从1开始
\newtheorem{lemma}{Lemma} % 定义引理环境

\newcommand{\halfmark}{\stackinset{c}{0pt}{c}{0pt}{\ding{55}}{\ding{51}}}
\definecolor{MyGray}{rgb}{0.85, 0.85, 0.85}

\definecolor{iccvblue}{rgb}{0.21,0.49,0.74}
\usepackage[pagebackref,breaklinks,colorlinks,allcolors=iccvblue]{hyperref}

\def\paperID{4659} % *** Enter the Paper ID here
\def\confName{ICCV}
\def\confYear{2025}

\title{SegmentDreamer: Towards High-fidelity Text-to-3D Synthesis with Segmented Consistency Trajectory Distillation}

\author{
    Jiahao Zhu\textsuperscript{1} \quad
    Zixuan Chen\textsuperscript{1} \quad
    Guangcong Wang\textsuperscript{2} \quad
    Xiaohua Xie\textsuperscript{1,3,4*} \quad
    Yi Zhou\textsuperscript{1*} \\
    \textsuperscript{1}Sun Yat-sen University \quad
    \textsuperscript{2}Great Bay University \quad
    \textsuperscript{3}Pazhou Lab (Huangpu) \\
    \textsuperscript{4}Guangdong Province Key Laboratory of Information Security Technology
     \\
    {\tt\small \{zhujh59,chenzx3\}@mail2.sysu.edu.cn,} 
    {\tt\small wanggc3@gmail.com,} 
    {\tt\small \{xiexiaoh6,zhouyi\}@mail.sysu.edu.cn}
}
\begin{document}
\twocolumn[{
\renewcommand\twocolumn[1][]{#1}%
\maketitle
\thispagestyle{empty}
\vspace{0.5em}
\begin{center}
    \centering
    \captionsetup{type=figure}
    \includegraphics[width=\textwidth]{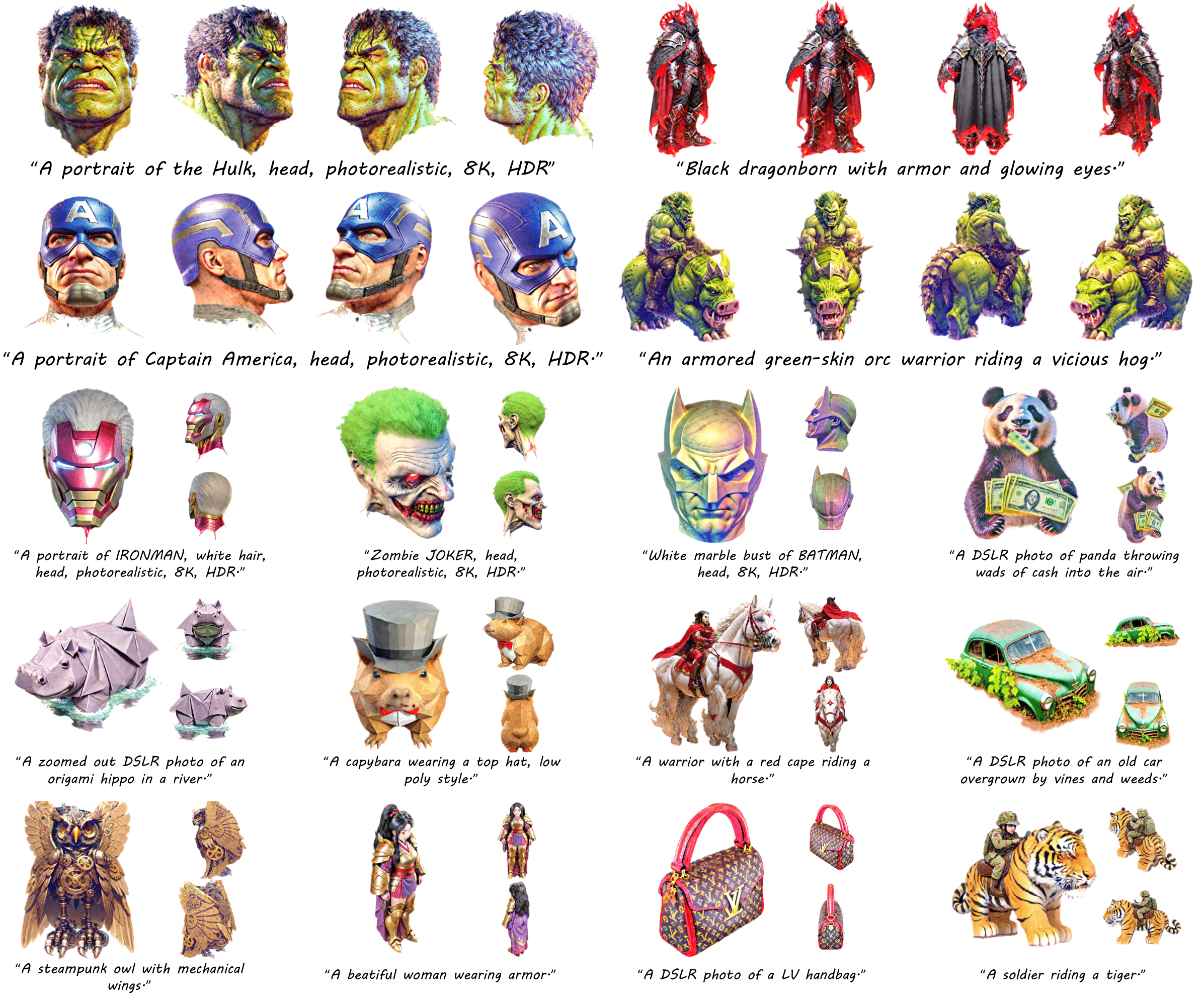}
    \captionof{figure}{
    \textbf{Examples of 3D assets generated by \textit{SegmentDreamer}.} Our framework addresses the improper conditional guidance issues of recent Consistency Distillation (CD)-based methods and theoretically provides a significantly tighter upper bound on distillation error, enabling high-fidelity text-to-3D generation within a reasonable timeframe ($\sim$32 minutes with classifier-free guidance \cite{cfg} and $\sim$38 minutes combined with Perp-Neg \cite{perp-neg}) on a single A100 GPU through 3D Gaussian Splatting (3DGS). Project Page: \href{https://zjhJOJO.github.io/segmentdreamer}{https://zjhJOJO.github.io/segmentdreamer}}
    \label{fig:fisrtvis}
\end{center}
}]
\renewcommand{\thefootnote}{}
\footnotetext{*Corresponding author.}
\renewcommand{\thefootnote}{\arabic{footnote}} % 恢复数字
\setcounter{footnote}{0} 
\begin{abstract}
% 3D assets can be distilled from text-to-image diffusion models. Recent advancements in Consistency Distillation (CD)-based methods significantly advance text-to-3D generation, addressing the issues of low fidelity in Score Distillation Sampling (SDS) and its variants. Nevertheless, these methods suffer from improper conditional guidance and less tight distillation error bound, leading to suboptimal generation. To overcome these limitations, we first reframe Classifier-free Guidance (CFG)-enhanced SDS loss from the perspective of CD, re-evaluating the role of CD in score distillation. Building on this reframing, we introduce a novel CD-based approach named Segmented Consistent Trajectory Distillation (SCTD). Unlike prior CD-based methods that enforce consistency across the entire ordinary differential equation trajectory, SCTD segments the trajectory, ensuring consistency within each sub-trajectory. This segmentation improves conditional guidance when using CFG and tightens the distillation error bound, leading to more accurate gradients for 3D generation. Extensive experiments demonstrate that our Gaussian Splatting based SegmentedDream surpasses state-of-the-art methods both in quality and training efficiency. Project is available at [link].

Recent advancements in text-to-3D generation improve the visual quality of Score Distillation Sampling (SDS) and its variants by directly connecting Consistency Distillation (CD) to score distillation.
However, due to the imbalance between self-consistency and cross-consistency, these CD-based methods inherently suffer from improper conditional guidance, leading to sub-optimal generation results.
To address this issue, we present \textbf{SegmentDreamer}, a novel framework designed to fully unleash the potential of consistency models for high-fidelity text-to-3D generation.
Specifically, we reformulate SDS through the proposed Segmented Consistency Trajectory Distillation (SCTD), effectively mitigating the imbalance issues by explicitly defining the relationship between self- and cross-consistency.
Moreover, \textbf{SCTD} partitions the Probability Flow Ordinary Differential Equation (PF-ODE) trajectory into multiple sub-trajectories and ensures consistency within each segment, which can theoretically provide a significantly tighter upper bound on distillation error.
Additionally, we propose a distillation pipeline for a more swift and stable generation.
Extensive experiments demonstrate that our \textbf{SegmentDreamer} outperforms state-of-the-art methods in visual quality, enabling high-fidelity 3D asset creation through 3D Gaussian Splatting (3DGS).

\end{abstract}    
\section{Introduction}
\label{sec:intro}
Text-to-3D generation aims to create 3D assets that accurately align with text descriptions.
It plays a vital role in various domains such as virtual reality, 3D gaming, and film production.
One compelling solution is to directly ``distill'' 3D assets from 2D text-to-image Diffusion Models (DM) \cite{stablediffusion} through Score Distillation Sampling (SDS) \cite{dreamfusion,wang2023score}, which leverages pre-trained text-to-image DMs as the 3D priors to guide the optimization of differentiable 3D representations, such as Neural Radiance Fields (NeRF) \cite{nerf} and 3D Gaussian Splatting (3DGS) \cite{3DGS}.
Subsequent research focuses on improving the performance of SDS in diversity \cite{prolificdreamer,dreamtime,diverseeccv}, visual quality \cite{fantasia3d,magic3d,prolificdreamer,dreamreward,vividdreamer,consistent3d,connectingconsistency,unidream,csd,hifa,nfsd}, and multi-view consistency \cite{textto3DGS,efficientdreamer,mvdream,let,jointdreamer}.

\begin{figure}[t]
  \centering
   %\includegraphics[width=8.6cm,height=4.2cm]{fig/framework.pdf}
   \includegraphics[width=\linewidth]{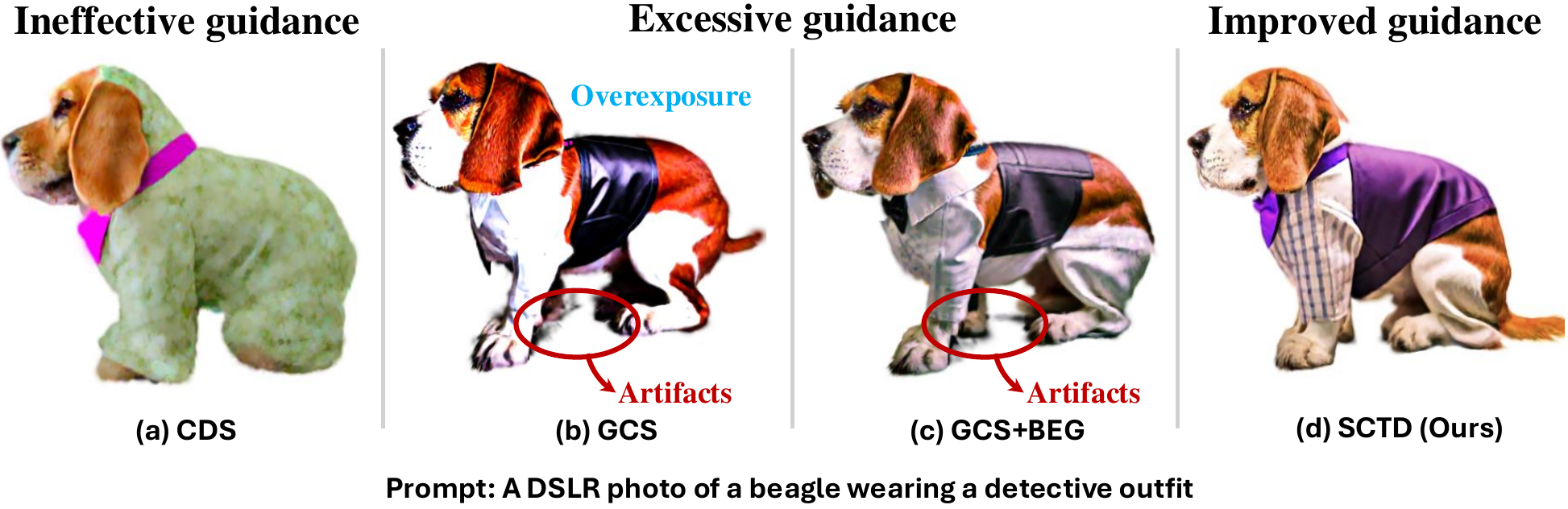}
   \caption{Visual examples of existing CD-based methods: CDS \cite{consistent3d}, GCS \cite{connectingconsistency}, and our \textit{SegmentDreamer} with a Classifier-Free Guidance (CFG) scale of 7.5.
   As shown, improper conditional guidance leads to suboptimal results.
   \textbf{(a)} Ineffective guidance produces semantically inconsistent details, while \textbf{(b)} excessive guidance results in overexposure and artifacts.
   Although \textbf{(c)} Brightness-Equalized Generation (BEG) \cite{connectingconsistency} can mitigate overexposure, artifacts persist.
   In contrast, our \textit{SegmentDreamer} enhances conditional generation, producing high-fidelity 3D assets.}
   \label{fig:example} 
\end{figure}

Recent methods \cite{consistent3d,connectingconsistency} have improved SDS and its variants by incorporating Consistency Distillation (CD) \cite{cm,ctm} into their distillation loss.
However, due to an imbalance between self-consistency and cross-consistency in their losses, these methods inherently suffer from improper conditional guidance when combined with Classifier-Free Guidance (CFG) for generation, ultimately leading to suboptimal outcomes (see \Cref{fig:example}; a detailed explanation is provided in \Cref{secsec:revisting}).
Additionally, the distillation error upper bound of these methods remains large, limiting their generation quality.

To address the aforementioned limitations, we present a novel framework -- \textit{SegmentDreamer}, which aims to fully unleash the potential of consistency models for high-fidelity text-to-3D generation.
Specifically, by analyzing existing CD-based methods, we draw inspiration from segmented consistency models \cite{hypersd,pcm} to reformulate SDS through the proposed Segmented Consistency Trajectory Distillation (SCTD), which effectively mitigates the imbalance between self-consistency and cross-consistency by explicitly defining their relationship.
Moreover, unlike CD-based methods that directly enforce consistency across the entire Probability Flow Ordinary Differential Equation (PF-ODE) trajectory, \textit{SCTD} partitions the PF-ODE trajectory into multiple sub-trajectories, ensuring consistency within each segment.
This approach theoretically provides a significantly tighter upper bound on distillation error.
In addition, we propose a distillation pipeline for a more swift and stable generation.
We also introduce 3DGS as our 3D representation.
Extensive experimental results demonstrate that our \textit{SegmentDreamer} outperforms state-of-the-art methods in visual quality.
Moreover, the proposed \textit{SCTD} can be seamlessly applied to various 3D generative tasks (\textit{e.g.}, 3D avatar generation and 3D portrait generation), facilitating the creation of high-fidelity 3D assets.

The main contributions are summarized as follows:
\begin{itemize}
    \item We present \textit{SegmentDreamer}, a novel framework for high-fidelity text-to-3D generation, which addresses the imbalance issue of existing CD-based methods.
    % We provide a comprehensive analysis of existing Consistency Distillation (CD)-based methods, identifying the underlying causes of improper conditional guidance in their distillation losses. 
    \item We reformulate SDS through the proposed \textit{SCTD}, effectively mitigating the imbalance between self- and cross-consistency.
    \textit{SCTD} also theoretically provides a significantly tighter upper bound on distillation error.
    We further propose a distillation pipeline for more swift and stable optimization. 
    % We introduce the Segmented Consistency Trajectory Distillation (SCTD) loss, which improves the conditional guidance and offers a tighter theoretical upper bound on distillation error, providing theoretical support for our generated results.
    \item Experimental results demonstrate that our \textit{SegmentDreamer} outperforms state-of-the-art methods in visual quality, enabling high-fidelity 3D asset generation through 3DGS.
\end{itemize}

\section{Related Work}
\label{sec:related_work}
\subsection{Text-to-3D Generation}
Creating 3D objects from text descriptions is challenging in the fields of computer vision and graphics.
To achieve this goal, a series of end-to-end text-to-3D pipelines have been proposed \cite{3dgen,get3d,pointe,shape-e,diffusionsdf,xiang2024structured}, but these methods require large-scale labeled 3D datasets for training.
To reduce the reliance on 3D training data, a compelling solution is to directly ``distill'' 3D assets from 2D text-to-image Diffusion Models (DM) \cite{stablediffusion}.
A groundbreaking contribution in this domain is DreamFusion \cite{dreamfusion}, which introduced the Score Distillation Sampling (SDS) loss to optimize parametric 3D representations.
Follow-up works focus on refining and extending SDS in various aspects.
For example, several studies \cite{prolificdreamer,dreamtime,diverseeccv} improve SDS in diversity through prompting modifications, while others \cite{fantasia3d,magic3d,prolificdreamer,dreamreward,vividdreamer,consistent3d,connectingconsistency,unidream,csd,hifa,nfsd} enhance the fidelity of the generated 3D objects.
Other advancements aim to achieve multi-view consistency \cite{textto3DGS,efficientdreamer,mvdream,let,jointdreamer} and faster generation \cite{dreamgaussian,gaussiandreamer}.
Recently, Consistency Distillation (CD) \cite{cm,ctm} has received widespread attention as it enables high-quality generation through minimal sampling steps.
Inspired by CD, Wu \textit{et al.} \cite{consistent3d} and Li \textit{et al.} \cite{connectingconsistency} incorporate CD into their distillation loss, which addresses some limitations brought by SDS.
However, these CD-based methods inherently suffer from improper conditional guidance, leading to sub-optimal results.
In this work, We aim to unleash CD's potential in generating high-fidelity 3D assets and push text-to-3D synthesis boundaries.

\subsection{Differentiable 3D Representations}
\noindent\textbf{Neural Radiance Field (NeRF)} \cite{nerf} aims to construct implicit 3D representations using a multi-layer perceptron (MLP), leveraging standard volumetric rendering \cite{rendering} and alpha compositing techniques \cite{alpha}.
Subsequent research has explored adapting NeRF to various applications, including sparse-view reconstruction \cite{Sparsenerf}, anti-aliasing \cite{mip-nerf}, and medical image super-resolution \cite{cunerf}.

\noindent\textbf{3D Gaussian Splatting (3DGS)} \cite{3DGS} has emerged as a leading approach for 3D representation, offering rapid optimization and high-speed rendering.
Unlike NeRF, 3DGS explicitly models scenes using a set of 3D Gaussians and renders views through splatting \cite{splatting}.
Its versatility has enabled applications across various domains, including avatar generation \cite{shao2024splattingavatar,yuan2024gavatar}, single-view reconstruction \cite{szymanowicz2024splatter}, anti-aliasing \cite{yu2024mip}, digital watermarking \cite{GuardSplat}, and SLAM \cite{matsuki2024gaussian,yan2024gs}.

% Neural Radiance Field (NeRF) \cite{nerf} is a classical view synthesis technique that optimizes an implicit volumetric scene function using a sparse collection of input views. Recently, NeRF, as a differentiable 3D representation, has gained significant attention in various 3D generation tasks \cite{dreamfusion,Zero123,haque2023instruct}. Despite its widespread use, optimizing NeRF can be computationally expensive and time-consuming. To address this, an alternative approach, 3D Gaussian Splatting (3DGS) \cite{3DGS}, has been introduced, significantly accelerating 3D generation applications and making them more practical. 3DGS employs an explicit scene representation, modeling a scene with millions of learnable 3D Gaussians in space. Considering 3DGS’s efficiency and its comparable performance to NeRF, we adopt it as our 3D representation for text-to-3D generation.
\section{Preliminaries}
\label{sec:preliminary}
\noindent\textbf{Diffusion Model (DM)} \cite{score-basd,ddim,karras} consists of a diffusion process and a denoising process. 
The diffusion process can be modeled as the solution of an It$\hat{\text{o}}$ stochastic differential equation (SDE) with the transition kernel $\mathbb{P}(\mathbf{z}_t|\mathbf{z}_0):=\mathcal{N}(\mathbf{z}_t;\alpha_t\mathbf{z}_0,\sigma_t^2\mathbf{I})$. 
By reversing the diffusion process, $\mathbf{z}_0$ can be estimated by solving a Probability Flow Ordinary Differentiable Equation (PF-ODE), which is defined by $\mathrm{d}\mathbf{z}_t = [f(t)\mathbf{z}_t-\frac{1}{2}g(t)^2\nabla_{\mathbf{z}_t}\mathrm{log}\mathbb{P}(\mathbf{z}_t)]\mathrm{d}t$. 
A neural network ${\boldsymbol{\epsilon_\phi}}(\mathbf{z},t)$ can be trained to approximate the scaled score function $-\sigma_t\nabla_{\mathbf{z}_t}\mathrm{log}\mathbb{P}(\mathbf{z}_t)$, offering an empirical formulation of the PF-ODE: $\mathrm{d}\mathbf{z}_t = [f(t)\mathbf{z}_t+\frac{g(t)^2}{2\sigma_t}\boldsymbol{\epsilon_\phi}(\mathbf{z}_t,t)]\mathrm{d}t.$
  
\noindent\textbf{Consistency Model} \cite{cm, lcm} aims to achieve 
few-step inference by distilling a pre-trained DM $\boldsymbol{\phi}$ into a student model $\boldsymbol{\phi}_{stu}$. Given a PF-ODE trajectory $\{\mathbf{z}_{t}\}_{t\in [0,T]}$ and a condition $\mathbf{y}$, a consistency model $\boldsymbol{f}_{\boldsymbol{\phi}_{stu}}$ builds the mapping \textbf{}$\boldsymbol{f}_{\boldsymbol{\phi}_{stu}}(\mathbf{z}_{t}, t, \mathbf{y})=\mathbf{z}_0$ between any points along this trajectory and the endpoint $\mathbf{z}_0$ by minimizing the following Consistency Distillation (CD) loss:
\begin{equation}
    \label{eq:cdloss}
    \mathcal{L}_{\text{CD}}=\mathbb{E}[\lambda(t)||\boldsymbol{f}_{\boldsymbol{\phi}_{stu}}(\mathbf{z}_{t},t),\boldsymbol{f}_{\boldsymbol{\phi}_{stu}^-}(\hat{\mathbf{z}}^{\boldsymbol{\Phi}}_{s},s)||_2^2], t>s,
\end{equation}
where 
$\mathbf{z}_{t}=\alpha_{t}\mathbf{z}_0+\sigma_{t}\boldsymbol{\epsilon}$ for $\boldsymbol{\epsilon}\sim \mathcal{N}(\mathbf{0},\mathbf{I})$ and $\boldsymbol{\phi}_{stu}^-$ is updated with the exponential moving average \cite{cm,lcm}. $\hat{\mathbf{z}}^{\boldsymbol{\Phi}}_{s}$ can be obtained by:
\begin{equation}
\begin{aligned}
    \hat{\mathbf{z}}^{\boldsymbol{\Phi}}_{s}&=\boldsymbol{\Phi}(\mathbf{z}_{t},t,s,\mathbf{y})\\
    &=\mathbf{z}_{t} + \int_{t}^{s}[f(u)\mathbf{z}_u+\frac{g(u)^2}{2\sigma_u}\boldsymbol{\epsilon_\phi}(\mathbf{z}_u,u,\mathbf{y})]\mathrm{d}u.
\end{aligned}
\end{equation}
The consistency function $\boldsymbol{f}_{\boldsymbol{\phi}_{stu}}$ can be parameterized in various ways \cite{cm,lcm}, and the solver $\boldsymbol{\Phi}(\cdot,\cdot,\cdot,\cdot)$ is used to calculate the point at $s$, which can be implemented using DDIM \cite{ddim} or DPM-solver \cite{dpm}.

% or Heun solver \cite{karras} in practice.
% A simple way to parameterize $\boldsymbol{f_\psi}$ is
% \begin{equation}
% \label{eq:cm_p}
%   \begin{aligned}
%     \boldsymbol{f_\psi}(\mathbf{z}_t,t)=
%     \begin{cases} 
%       \mathbf{z}_\epsilon, & \text{if } t = \epsilon, \\ 
%       \boldsymbol{F_\psi}(\mathbf{z}_t,t), & \text{if } t \neq \epsilon.
%       \end{cases}
%   \end{aligned}
% \end{equation}
% Here $\boldsymbol{F_\psi}$ denotes the single-step prediction of $\mathbf{z}_\epsilon$ using ODE solvers like DDIM \cite{ddim} or DPM-solver \cite{dpm}.

\noindent\textbf{Score Distillation Sampling (SDS) Loss}. Given a 3D representation $\boldsymbol{\theta}$, one can render a view image $\mathbf{z}_0:=\boldsymbol{g}(\mathbf{c},\boldsymbol{\theta})$ with camera pose $\mathbf{c}$, SDS distills a 3D object from a 2D DM $\boldsymbol{\phi}$ by minimizing the following SDS loss \emph{w.r.t.} $\boldsymbol{\theta}$:
\begin{equation}
\label{eq:sdsloss}
    \mathcal{L}_{\text{SDS}}:=\mathbb{E}_{t,\boldsymbol{\epsilon}}[\omega(t)||\underbrace{\boldsymbol{\epsilon_\phi}(\mathbf{z}_t,t,\mathbf{y})-\boldsymbol{\epsilon}}_{\text{noise residual}}||_2^2],
\end{equation}
where $\mathbf{z}_t=\alpha_t\mathbf{z}_0+\sigma_t\boldsymbol{\epsilon}$.
Poole \emph{et al.} \cite{dreamfusion} omit the Jacobian term $\frac{\partial \boldsymbol{\epsilon_\phi}(\mathbf{z}_t,t,\mathbf{y})}{\partial\mathbf{z}_t}$ in gradient calculation for speedup.

% When computing the gradient of $\mathcal{L}_{\text{SDS}}$ \emph{w.r.t.} $\boldsymbol{\theta}$, Poole \emph{et al.} \cite{dreamfusion} empirically observed that omitting the U-Net Jacobian term, \emph{i.e.}, $\frac{\partial \boldsymbol{\epsilon_\phi}(\mathbf{z}_t,t,\mathbf{y})}{\mathbf{z}_t}$, also results in an effective gradient for optimization.
% \begin{equation}
% \label{eq:sdsloss}
%     \nabla_{\boldsymbol{\eta}}\mathcal{L}_{\text{SDS}} :=\mathbb{E}_{t,\boldsymbol{\epsilon}}[\omega(t)(\boldsymbol{\epsilon_\phi}(\mathbf{z}_t,t,\mathbf{y})-\boldsymbol{\epsilon})\frac{\partial \mathbf{z_c}}{\partial \boldsymbol{\theta}}].
% \end{equation}
\section{Our Method}
\label{sec:ourmethod}
\begin{figure*}[!t]
  \centering
   \includegraphics[width=0.95\linewidth, height=6cm]{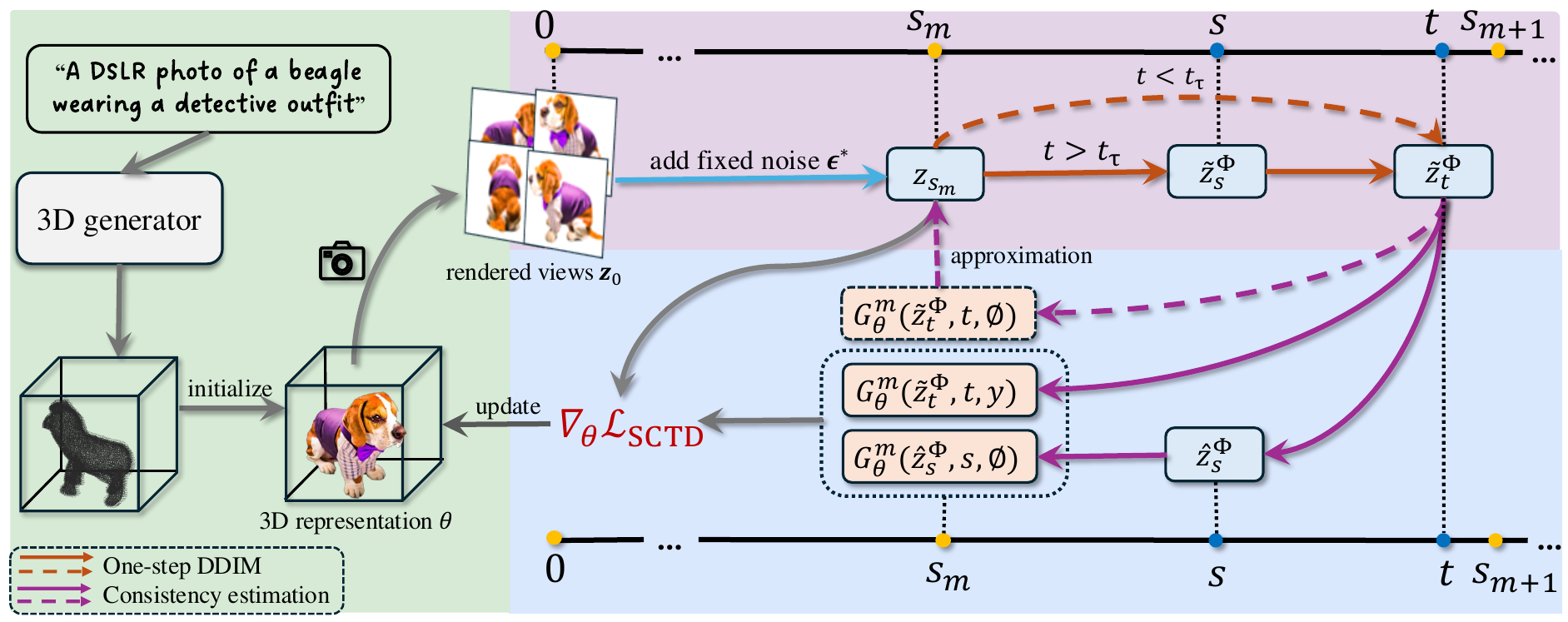}
   \caption{An overview of \textbf{SegmentDreamer}: We begin by initializing a 3D representation $\boldsymbol{\theta}$ using a 3D generator, such as Point-E~\cite{pointe}. In each iteration, we randomly render a batch of camera views $\mathbf{z}_0$ from $\boldsymbol{\theta}$ and diffuse them into $\mathbf{z}_{s_m}$ with fixed noise $\boldsymbol{\epsilon}^*$. Next, we transform $\mathbf{z}_{s_m}$ into $\tilde{\mathbf{z}}^{\boldsymbol{\Phi}}_t$ using either one-step or two-step unconditional deterministic sampling. During the denoising process, we first estimate $\hat{\mathbf{z}}^{\boldsymbol{\Phi}}_s$ through one-step conditional deterministic sampling from $\tilde{\mathbf{z}}^{\boldsymbol{\Phi}}_t$. Subsequently, we compute two consistency functions and utilize them to derive the loss $\mathcal{L}_{\text{SCTD}}$, which is ultimately employed to optimize $\boldsymbol{\theta}$.}
    \label{fig:SCTD_framework}
\end{figure*}

In this section, we first review the mechanism of existing Consistency Distillation (CD)-based methods \cite{consistent3d,connectingconsistency} and reveal that they inherently suffer from improper conditional guidance when combined with Classifier-free Guidance (CFG).
Subsequently, we address these limitations by reformulating SDS through the lens of the proposed Segmented Consistency Trajectory Distillation (SCTD).
Moreover, building on this insight, we propose an efficient SCTD sampling method for high-fidelity text-to-3D generation.
We further tailor a distillation pipeline for the proposed SCTD sampling to achieve more swift and stable optimization.
Finally, we provide a theoretical proof for our distillation error upper bound.
The overall framework of our \textit{SegmentDreamer} is depicted in \cref{fig:SCTD_framework}, with the algorithm detailed in \cref{alg:SCTD}.

% In \cref{secsec:revisting}, we analyze previous Consistency Distillation (CD)-based methods \cite{consistent3d,connectingconsistency}, identifying potential causes of their limitations. In \cref{secsec:cd-sds}, we address these limitations by reframing the Classifier-free Guidance (CFG)-based Score Distillation Sampling (SDS) loss through the lens of Segmented Consistency Trajectory Distillation (SCTD). Building on this insight, in \cref{secsec:tscd_loss}, we introduce the efficient SCTD sampling and provide theoretical proof for our distillation error upper bound. The overall framework of SegmentDreamer is illustrated in \cref{fig:SCTD_framework}, with the algorithm detailed in \cref{alg:SCTD}.

\subsection{Reviewing CD-based Losses}
The consistency function \textit{w.r.t.} the 3D representation $\boldsymbol{\theta}$ is parameterized as $\boldsymbol{F_\theta}(\mathbf{z}_{t},t,\mathbf{y})= \frac{\mathbf{z}_{t}-\sigma_t\boldsymbol{\epsilon_\phi}(\mathbf{z}_{t},t,\mathbf{y})}{\alpha_t}$ \cite{consistent3d,connectingconsistency}. 
Before proceeding, we clarify two key concepts: \emph{self-consistency} and \emph{cross-consistency}. 
The former ensures that all points on the same ODE trajectory consistently map to their original points, as in consistency models \cite{lcm,cm,ctm}. The latter enforces alignment between unconditional and conditional ODE trajectories \cite{connectingconsistency}.
\label{secsec:revisting}

\noindent \textbf{Consistency Distillation Sampling (CDS)} loss \cite{consistent3d} is defined as
\begin{equation}
\label{eq:cdsloss}
\begin{aligned}
        \mathcal{L}_{\text{CDS}}(\boldsymbol{\theta}) =\mathbb{E}_{t,s}[c(t)||\boldsymbol{F_\theta}(\mathbf{z}_{t},t,\mathbf{y})-\boldsymbol{F_\theta}(\hat{\mathbf{z}}^{\boldsymbol{\Phi}}_{s},s,\mathbf{y})||_2^2],
\end{aligned}
\end{equation}
where $s<t$, $c(t)=\omega(t)(\frac{\alpha_{t}}{\sigma_{t}})^2$, and $\hat{\mathbf{z}}^{\boldsymbol{\Phi}}_{s}$ is estimated using single-step DDIM sampling.
\cref{eq:cdsloss} reveals that CDS prioritizes self-consistency in the whole ODE trajectory while omitting cross-consistency in its loss function, which can be interpreted as the absence of the ``classifier score''.
As reported in CSD \cite{csd}, this missing component, which is crucial for conditional generation, hinders CDS from producing semantically consistent content (see \cref{fig:loss_compare} for details).
Moreover, the reliance on  $\boldsymbol{F}_\theta$ may introduce a significant distillation error upper bound $\mathcal{O}(\Delta_t)T$, where $\Delta_t = \max{|t - s|}$.
This issue is particularly pronounced in text-to-3D settings due to the typically large $|t-s|$.

% In other words, CDS lacks the ``classifier score'', a crucial component for conditional generation, as demonstrated in Classifier Score Distillation (CSD) \cite{csd}. Consequently, CDS struggles to generate effectively under normal CFG scales, as further evidenced in \cref{fig:loss_compare}. 

\noindent \textbf{Guided Consistency Sampling (GCS)} \cite{connectingconsistency} leverages the DPM-Solver-1 \cite{dpm} as its consistency function:
\begin{equation}
  \label{eq:gcs_cf}
    \boldsymbol{G_\theta}(\mathbf{z}_{t}, t, s,\mathbf{y})=\frac{\alpha_s}{\alpha_t}\mathbf{z}_{t}-\alpha_s\boldsymbol{\epsilon}_{\boldsymbol{\phi}}(\mathbf{z}_{t},t,\mathbf{y})\int_{\lambda_t}^{\lambda_s}e^{-\lambda}\mathrm{d}\lambda,
\end{equation}
where $\lambda_t=\mathrm{ln}\frac{\alpha_t}{\sigma_t}$.
Based on \cref{eq:gcs_cf}, GCS proposes a more compact consistency (CC) loss $\mathcal{L}_{\text{CC}}$ defined as
\begin{equation}
    \label{eq:ccloss}
    \mathcal{L}_{\text{CC}}(\boldsymbol{\theta})\!=\! \mathbb{E}_{t,s,e}[\left\| \boldsymbol{G_\theta}(\tilde{\mathbf{z}}^{\boldsymbol{\Phi}}_{t}, t, e, \emptyset)\!-\! \boldsymbol{G_\theta}(\hat{\mathbf{z}}_s^{\boldsymbol{\Phi}},s, e, \emptyset) \right\|_2^2],
\end{equation}
where $e$$<$$s$$<$$t$. GCS also employs a conditional guidance loss $\mathcal{L}_{\text{CG}}$ and a pixel-wise constraint $\mathcal{L}_{\text{CP}}$ to enforce cross-consistency (see Supp. \ref{app:gcs} for details).
%From \cref{eq:cdloss}, it can be derived that its distillation error upper bound is $\mathcal{O}(\Delta_t)(T-e)$, where $\Delta_t=max\{t-s\}$. 

GCS ensures self-consistency within sub-trajectory, but it still suffers from improper conditional guidance due to two key issues: (1) $\boldsymbol{G_\theta}$ used to enforce cross-consistency is inherently flawed because $\boldsymbol{\epsilon}_{\boldsymbol{\phi}}(\mathbf{z}_{t},t,\mathbf{y})$ lacks a target time-step, as detailed in App. \ref{app:flaw}. (2) GCS employs $\boldsymbol{F}_{\boldsymbol{\theta}}$ (implemented via one-step DDIM) to enforce cross-consistency throughout the entire ODE trajectory in both latent and pixel spaces. 
However, as reported in its paper, it suffers from unstable generation due to the excessive conditional guidance, leading to overexposure and artifacts (more experimental analysis is provided in App. \cref{app:oversaturation}).
Furthermore, this approach results in a less stringent upper bound on the distillation error, expressed as $\mathcal{O}(\Delta_t)(T-e)$, which constrains its generation quality.

% Although a remedial approach, Brightness-Equalized Generation (BEG), has been proposed, the problems of artifacts and slow training persist, as evidenced in \cref{fig:loss_compare}.
% and $\tilde{\mathbf{z}}^{\boldsymbol{\Phi}}_{t}$ is estimated by the following trajectory: $\tilde{\mathbf{z}}^{\boldsymbol{\Phi}}_{s}=\boldsymbol{\Phi}(\mathbf{z}_{e},e, s,\emptyset)\to\tilde{\mathbf{z}}^{\boldsymbol{\Phi}}_{t}=\boldsymbol{\Phi}(\tilde{\mathbf{z}}^{\boldsymbol{\Phi}}_{s},s, t,\emptyset)$; and $\hat{\mathbf{z}}^{\boldsymbol{\Phi}}_{s}=\boldsymbol{\Phi}(\tilde{\mathbf{z}}^{\boldsymbol{\Phi}}_{t}, t, s, \mathbf{y})$. 
\begin{table}[!t]
\caption{Differences between CDS, GCS, and SCTD.``traj.'' denotes trajectory. }
\setlength{\tabcolsep}{1.3mm}
\centering
\scriptsize
\begin{tabular}{c|ccc}
\toprule
 & \textbf{CDS \cite{consistent3d}} & \textbf{GCS} \cite{connectingconsistency} & SCTD (Ours) \\ \hline
% \textbf{CFG scale $\omega$} & 50+ & 7.5+ & 7.5+ \\ 
% \textbf{Consistency function} & $\boldsymbol{F}_{\boldsymbol{\theta}}(\mathbf{z}_t,t,\mathbf{y})$  & $\boldsymbol{G}_{\boldsymbol{\theta}}(\mathbf{z}_t,t,s,\mathbf{y})$ & $\boldsymbol{G}^m_{\boldsymbol{\theta}}(\mathbf{z}_t,t,\mathbf{y})$ \\
% \textbf{Objective function} & $\mathcal{L}_{\text{CDS}}$ & $\mathcal{L}_{\text{CC}}+\mathcal{L}_{\text{CG}}+\mathcal{L}_{\text{CP}}$ & $\mathcal{L}_{\text{SCTD}}$ \\ 
\textbf{Self-consistency} & 
\ding{51}(whole \textit{traj.}) & \ding{51} (sub \textit{traj.}) & \ding{51} (sub \textit{traj.})\\
\textbf{Cross-consistency} & \ding{55} & \halfmark (whole \textit{traj.}) & \ding{51}(sub \textit{traj.})\\
\textbf{Error bound} & $\mathcal{O}(\Delta_t)T$ & $\mathcal{O}(\Delta_t)(T\!-\!e)$ & $\mathcal{O}(\Delta_t)(s_{m+1}\!-\!s_m)$\\
\bottomrule
\end{tabular}
\label{rebuttal:comparison}
\end{table}
\noindent \textbf{Summary:} The improper conditional guidance in CDS and GCS stems mainly from an imbalance between self-consistency and cross-consistency. Additionally, their reliance on $\boldsymbol{F_\theta}$ to enforce consistency across the entire trajectory exacerbates this issue. \cref{rebuttal:comparison} clarifies the differences between CDS, GCS, and ours.
\subsection{Reframing SDS through SCTD}
We address the previous limitations by reframing SDS through the lens of SCTD.
Specifically, we first define our consistency function as follows: as existing segmented consistency models \cite{pcm,hypersd} do, we partition the entire time-step range $[0,T]$ into $N_s$ sub-intervals (\emph{i.e.}, $N_s+1$ edge points), represented by $\{[s_m,s_{m+1})|m\in\{0,\cdots,N_s-1\}, s_0=0, s_{N_s}=T\}$ (the segmentation strategy is discussed in \cref{secsec:tscd_loss}). We define our consistency function $\boldsymbol{G_\theta}$ using \cref{eq:gcs_cf}, aiming to enforce consistency within each sub-trajectory, \emph{i.e.}, $\boldsymbol{G}_{\boldsymbol{\theta}}(\mathbf{z}_{t},t,s_m,\mathbf{y})=\boldsymbol{G}_{\boldsymbol{\theta}}(\mathbf{z}_{s},s,s_m,\mathbf{y})$ for $s_m\leq s<t\leq s_{m+1}$. For brevity, we abbreviate $\boldsymbol{G}_{\boldsymbol{\theta}}(\mathbf{z}_{t},t,s_m,\mathbf{y})$ as $\boldsymbol{G}_{\boldsymbol{\theta}}^m(\mathbf{z}_{t},t,\mathbf{y})$. Notably, our consistency function $\boldsymbol{G}_{\boldsymbol{\theta}}^m$ avoids the issues inherent in GCS, as analyzed in \cref{secsec:revisting} and Supp. \ref{app:flaw}, because at each training stage, the target time step $s_m$ is always uniquely determined.

\label{secsec:cd-sds}
When CFG \cite{cfg} with scale $\omega$ is used, the noise residual term in \cref{eq:sdsloss} can be transformed into
\begin{equation}
\label{eq:sds-cfg}
\begin{aligned}
    \hat{\boldsymbol{\epsilon}}_{\boldsymbol{\phi}}(\mathbf{z}_{t},t,\mathbf{y})-\boldsymbol{\epsilon}&=\boldsymbol{\epsilon_\phi}(\mathbf{z}_{t},t,\mathbf{y})-\boldsymbol{\epsilon}\\
    &+\omega(\boldsymbol{\epsilon_\phi}(\mathbf{z}_{t},t,\mathbf{y})-\boldsymbol{\epsilon_\phi}(\mathbf{z}_{t},t,\emptyset)).\\
\end{aligned}
\end{equation}
Based on \cref{eq:sds-cfg}, $\mathcal{L}_{\text{SDS}}$ can be expressed in the form of SCTD: 
\begin{equation}
    \begin{aligned}
      & \mathcal{L}_{\text{SDS}}(\boldsymbol{\theta})=\mathbb{E}_{t,s}[b(t)||\underbrace{\boldsymbol{G}^m_{\boldsymbol{\theta}}(\hat{\mathbf{z}}_{s}^{\boldsymbol{\Phi}},s,\mathbf{y})-\boldsymbol{G}^m_{\boldsymbol{\theta}}(\mathbf{z}_{t},t,\mathbf{y})}_{\text{self-consistency constraint}}+\\
      &\omega\underbrace{(\boldsymbol{G}^m_{\boldsymbol{\theta}}(\mathbf{z}_{t},t, \emptyset)\!-\!\boldsymbol{G}^m_{\boldsymbol{\theta}}(\mathbf{z}_{t},t,\mathbf{y}))}_{\text{cross-consistency constraint}}
      \!+\!\underbrace{\mathbf{z}_{s_m} -\boldsymbol{G}^m_{\boldsymbol{\theta}}(\hat{\mathbf{z}}_{s}^{\boldsymbol{\Phi}},s,\mathbf{y})}_{\text{generative prior}}||_2^2]. 
    \end{aligned}
    \label{eq:noise-cd-tscd}
  \end{equation}
where $b(t)=\frac{\omega(t)}{(\alpha_{s_m}\int_{\lambda_{t}}^{\lambda_{s_m}}e^{-\lambda}\mathrm{d}\lambda)^2}$ and $\hat{\mathbf{z}}_{s}^{\boldsymbol{\Phi}}=\boldsymbol{\Phi}(\mathbf{z}_t,t,s,\mathbf{y})$. The derivation is given in App. \ref{app:proof4SCTD}.
The first term on the r.h.s. of \cref{eq:noise-cd-tscd} enforces self-consistency across the conditional denoising sub-trajectory, referred to as the "self-consistency constraint." The second term enforces cross-consistency between the unconditional and conditional denoising sub-trajectories, termed the "cross-consistency constraint." The third term adheres to the naming conventions in CSD \cite{csd}. 
As observed, \cref{eq:noise-cd-tscd} explicitly defines the relationship between the self-consistency constraint and the cross-consistency constraint. 
Furthermore, as demonstrated in \cref{secsec:tscd_loss}, SCTD provides a tighter upper bound on the distillation error.
% Notably, prior work \cite{csd} suggests discarding both the CD loss and the generator loss from \cref{eq:noise-cd}, arguing that the classifier loss plays a dominant role in optimization. However, we assert that removing the CD loss is unjustified. On one hand, the CD loss provides a theoretical upper bound for the distillation error. On the other hand, experimental results show that omitting the CD loss leads to suboptimal generation quality, as shown in \cref{sec:ablationstudy}.
\subsection{SCTD Sampling}
\label{secsec:tscd_loss}
\noindent \textbf{Sampling Design}. A natural question arises: according to \cref{eq:noise-cd-tscd}, does minimizing $\mathcal{L}_{SDS}$ necessarily enforce both self- and cross-consistency? The answer is no. This is because one cannot necessarily achieve $x_1 = x_2$ and $y_1 = y_2$ when $||x_1 - x_2 + \omega(y_1 - y_2)||_2^2=0$. 
% While \cref{eq:noise-cd-tscd} provides valuable insights, applying it to text-to-3D tasks remains challenging. 
Additionally, we observe that the condition $\mathbf{y}$ appears in all three loss terms in \cref{eq:noise-cd-tscd}, which increases the computational cost, particularly when using Perp-Neg \cite{perp-neg} or ControlNet \cite{controlnet}. 

To improve the computational efficiency, 
% we introduce the following modifications. First, we observe that the condition $\mathbf{y}$ appears in all three loss terms in \cref{eq:noise-cd-tscd}, which increases the computational cost, particularly when using Perp-Neg \cite{perp-neg} or ControlNet \cite{controlnet}. To mitigate this, 
we perform the following equivalent transformation on \cref{eq:noise-cd-tscd}: 
\begin{equation}
  \label{eq:noise-cd-tscd-mod}
    \begin{aligned}
      & \mathcal{L}_{\text{SDS}}(\boldsymbol{\theta})=\mathbb{E}_{t,s}[b(t)||\underbrace{\boldsymbol{G}^m_{\boldsymbol{\theta}}(\hat{\mathbf{z}}_{s}^{\boldsymbol{\Phi}},s,\emptyset)-\boldsymbol{G}^m_{\boldsymbol{\theta}}(\mathbf{z}_{t},t,\emptyset)}_{\text{self-consistency constraint}}+\\
      &(\omega\!+\!1)\!\underbrace{(\boldsymbol{G}^m_{\boldsymbol{\theta}}(\mathbf{z}_{t},t, \emptyset)\!-\!\boldsymbol{G}^m_{\boldsymbol{\theta}}(\mathbf{z}_{t},t,\mathbf{y}))}_{\text{cross-consistency constraint}}
      \!+\!\underbrace{\mathbf{z}_{s_m}\!-\!\boldsymbol{G}^m_{\boldsymbol{\theta}}(\hat{\mathbf{z}}_{s}^{\boldsymbol{\Phi}},s,\emptyset)}_{\text{generative prior}}\!||_2^2]. 
    \end{aligned}
  \end{equation}
  The derivation is given in Supp. \ref{app:proof4crt}.
  
   To enforce self- and cross-consistency, we follow the suggestion from CSD \cite{csd} to omit the generative prior from \cref{eq:noise-cd-tscd-mod}, and then impose a stricter constraint on \cref{eq:noise-cd-tscd}:
  \begin{equation}
    \label{eq:noise-cd-tscd-sim}
      \begin{aligned}
        & \mathcal{L}_{\text{SCTD}}(\boldsymbol{\theta})=\mathbb{E}_{t,s}[b(t)[||\underbrace{sg(\boldsymbol{G}^m_{\boldsymbol{\theta}}(\hat{\mathbf{z}}_{s}^{\boldsymbol{\Phi}},s,\emptyset))-\boldsymbol{G}^m_{\boldsymbol{\theta}}(\tilde{\mathbf{z}}_{t}^{\boldsymbol{\Phi}},t,\emptyset)}_{\text{self-consistency constraint}}||_2^2\\
        &+||(\omega+1)\underbrace{(\boldsymbol{G}^m_{\boldsymbol{\theta}}(\tilde{\mathbf{z}}_{t}^{\boldsymbol{\Phi}},t, \emptyset)-sg(\boldsymbol{G}^m_{\boldsymbol{\theta}}(\tilde{\mathbf{z}}_{t}^{\boldsymbol{\Phi}},t,\mathbf{y})))}_{\text{cross-consistency constraint}}||_2^2]].
      \end{aligned}
    \end{equation}
where $\tilde{\mathbf{z}}_{t}^{\boldsymbol{\Phi}}$ is obtained through multi-step deterministic sampling from $\mathbf{z}_0$, as discussed in \cref{secsec:SSO}. $\mathbf{z}_{s_m}=\alpha_{s_m}\mathbf{z}_0+\sigma_{s_m}\boldsymbol{\epsilon}^*$, with $\boldsymbol{\epsilon}^*$ fixed throughout the entire training. $\hat{\mathbf{z}}_{s}^{\boldsymbol{\Phi}}=\boldsymbol{\Phi}(\tilde{\mathbf{z}}_{t}^{\boldsymbol{\Phi}},t,s,\mathbf{y})$, where $\boldsymbol{\Phi}$ is implemented by a first-order ODE solver in this work. $sg(\cdot)$ represents the stopping of gradient backpropagation.
% and the hyperparameter $t_\tau$ controls the number of deterministic sampling steps in the diffusion process of $\mathbf{z}_t$. 
\begin{algorithm}[!t]
  \caption{Text-to-3D Generation via SCTD Loss}
  \begin{algorithmic}[1]
  \State \textbf{Initialization:} threshold $t_\tau$, segmentation number $N_s$, and target prompt $\mathbf{y}$
  \State Initialize $\boldsymbol{\theta}$ with a pre-trained 3D generator
  \While{$\boldsymbol{\theta}$ is not converged}
      \State Sample $\mathbf{z}_0 = \boldsymbol{g}(\mathbf{c},\boldsymbol{\theta})$ and $t \sim \mathcal{U}(0, T)$
      \State Obtain $s_m \!=\! max\{s_i|s_i\leq t,i=0,\cdots,N_s-1\}$
      \State Obtain $\mathbf{z}_{s_m}=\alpha_{s_m}\mathbf{z}_0+\sigma_{s_m}\boldsymbol{\epsilon}^*$
      \State Sample $s \sim \mathcal{U}(s_m, t)$
      \State Predict $\tilde{\mathbf{z}}_{t}^{\boldsymbol{\Phi}}$ with \cref{eq:SCTD_ode_sampling} and $\hat{\mathbf{z}}_{s}^{\boldsymbol{\Phi}}\!=\!\boldsymbol{\Phi}(\tilde{\mathbf{z}}_{s}^{\boldsymbol{\Phi}},t,s,\mathbf{y})$
      \State Obtain $\boldsymbol{G}^m_{\boldsymbol{\theta}}(\hat{\mathbf{z}}_{s}^{\boldsymbol{\Phi}},s,\emptyset)$ and $\boldsymbol{G}^m_{\boldsymbol{\theta}}(\tilde{\mathbf{z}}_{t}^{\boldsymbol{\Phi}},t,\mathbf{y})$ with \cref{eq:gcs_cf}
      \State Obtain $\nabla_{\boldsymbol{\theta}} \mathcal{L}_{\text{SCTD}}$ with \cref{eq:noise-cd-tscd-sim} and update $\boldsymbol{\theta}$.
  \EndWhile
  \end{algorithmic}
  \label{alg:SCTD}
  \end{algorithm}
% As demonstrated in \cite{connectingconsistency,hifa}, the absence of pixelwise constraints can result in unrealistic colors in the decoded images, introducing undesirable artifacts in the generated 3D objects. Based on this, we integrate high-resolution image supervision into $\mathcal{L}_{\text{SCTD}}$, leading to the final objective function:
% \begin{equation}
%   \label{eq:SCTD-loss-total}
%   \begin{aligned}
%     &\mathcal{L}_{\text{SCTD}}^{total}(\boldsymbol{\theta})=\mathcal{L}_{\text{TSCD}}(\boldsymbol{\theta}) + \beta \mathcal{L}_{\text{CP}}(\boldsymbol{\theta}),\\
%     &\mathcal{L}_{CP}(\boldsymbol{\theta})=\mathbb{E}_{t}[||\boldsymbol{D_\phi}(\boldsymbol{F_\theta}(\boldsymbol{G}^m_{\boldsymbol{\theta}}(\tilde{\mathbf{z}}_{t}^{\boldsymbol{\Phi}},t,\mathbf{y}),s_m,\mathbf{y}))-\boldsymbol{D_\phi}(\mathbf{z}_{0}^{\mathbf{c}})||_2^2], \\
%   \end{aligned}
% \end{equation}
% where $\boldsymbol{D_\phi}$ denotes the VAE decoder of the pre-trained DM.

% \noindent \textbf{Prompt Enhancement}. When using Perp-Neg \cite{perp-neg} or ControlNet \cite{controlnet} to enhance generation quality, the prompt embeddings $\mathbf{y}$ in \cref{eq:SCTD-loss-total} can simply be replaced with the corresponding enhanced embeddings $\mathbf{y}^*$. Similarly, for negative prompt embeddings $\mathbf{y}_{neg}$, $\emptyset$ in \cref{eq:SCTD-loss-total} should be substituted with $\mathbf{y}_{neg}$.

\noindent \textbf{Trajectory Segmentation Strategy}. We propose two heuristic trajectory segmentation strategies: equal segmentation and monotonically increasing segmentation. In the first strategy, the entire time-step range $[0,T]$ is uniformly divided into $N_s$ segments,  \emph{i.e.}, $s_{m+1} - s_m = \frac{T}{N_s}$. In the second strategy, we adopt a monotonically increasing interval segmentation method, where the $N_s$ intervals satisfy that $s_{m+1} - s_m = t_\tau + \frac{2m(T - N_s t_\tau)}{N_s(N_s - 1)}$ for $\in\{0,\cdots,N_s-1\}$. A comparison of the two strategies is presented in \cref{sec:ablationstudy}.

\subsection{Swift and Stable Optimization Pipeline}
In previous work \cite{luciddreamer,consistent3d,connectingconsistency}, $\tilde{\mathbf{z}}_t^{\boldsymbol{\Phi}}$ is derived from deterministic sampling with a fixed number of steps, lacking flexibility. In this paper, we adopt the following dynamic adjustment strategy:
\noindent\textbf{}.
\label{secsec:SSO}
\begin{equation}
  \label{eq:SCTD_ode_sampling}
  \tilde{\mathbf{z}}_{t}^{\boldsymbol{\Phi}} =
  \begin{cases} 
    \boldsymbol{\Phi}(\boldsymbol{\Phi}(\mathbf{z}_{s_m},s_m,s,\emptyset),s_m,t,\emptyset), & \text{if } t > t_\tau \\ 
    \boldsymbol{\Phi}(\mathbf{z}_{s_m},s_m,t,\emptyset), & \text{if } t \leq t_\tau
  \end{cases},
\end{equation}
This strategy enables us to make a trade-off between generation quality and optimization speed, as evidenced in \cref{sec:ablationstudy}.

\noindent\textbf{Consistency function Approximation}. For faster optimization, we propose approximating $\boldsymbol{G}^m_{\boldsymbol{\theta}}(\tilde{\mathbf{z}}_{t}^{\boldsymbol{\Phi}},t, \emptyset)$ as $\mathbf{z}_{s_m}$, leveraging the theoretical invertibility of the unconditional PF-ODE trajectory defined in \cref{eq:SCTD_ode_sampling}. However, since both $\boldsymbol{\Phi}$ and $\boldsymbol{G}_{\boldsymbol{\theta}}^m$ are implemented using a first-order ODE solver in this paper, perfect invertibility cannot be guaranteed. Nevertheless, we find that this approximation strategy not only eliminates the need to compute the U-Net Jacobian—reducing computational cost—but also improves optimization stability, as demonstrated in \cref{sec:ablationstudy}. 

\subsection{Theoretical Analysis of Distillation Error}
We theoretically prove that our SCTD loss leads to a tighter upper bound on the distillation error, which explains why SCTD achieves superior generative results. Specifically, the SDS loss can be reformulated as:
  \begin{equation}
    \label{eq:sdslosswocfg}
    \begin{aligned}
      \mathcal{L}_{\text{SDS}}&=\mathbb{E}_{t}[\omega(t)||\boldsymbol{\epsilon_\phi}(\tilde{\mathbf{z}}_{t}^{\boldsymbol{\Phi}},t,\mathbf{y})-\boldsymbol{\epsilon}^*||_2^2]\\
      % &=\mathbb{E}_{t}[c(t)||\mathbf{z}_{0}-\boldsymbol{F_\theta}(\tilde{\mathbf{z}}_{t}^{\boldsymbol{\Phi}}, t,\mathbf{y})||_2^2]\\
      &=\mathbb{E}_{t}[b(t)||\mathbf{z}_{s_m}-\boldsymbol{G}_{\boldsymbol{\theta}}^m(\tilde{\mathbf{z}}_{t}^{\boldsymbol{\Phi}}, t,\mathbf{y})||_2^2].
    \end{aligned}
  \end{equation}
Based on \cref{eq:sdslosswocfg}, we have the following theorem:
\begin{theorem}
  \label{th:1}
  Given a sub-trajectory $[s_m,s_{m+1}]$, let $\Delta t=max_{t,s\in(s_m,s_{m+1}]}\{|t-s|\}$. We assume $\boldsymbol{G}^m_{\boldsymbol{\theta}}$ satisfies the Lipschitz condition. If there exists a $\mathbf{z}_0$ satisfying $\boldsymbol{G}^m_{\boldsymbol{\theta}}(\tilde{\mathbf{z}}_{t}^{\boldsymbol{\Phi}},t,\mathbf{y})=\boldsymbol{G}^m_{\boldsymbol{\theta}}(\hat{\mathbf{z}}_{s}^{\boldsymbol{\Phi}},s,\mathbf{y})$ for $\forall t, s\in[s_m,s_{m+1}]$ and a true image $\mathbf{z}^{data}\sim \mathbb{P}_{data}(\mathbf{z})$ (corresponding to $\mathbf{z}_0$), we have 
  % \begin{equation}
  %     \begin{aligned}
  %   &\sup_{t,s\in[s_m,s_{m+1}]}||\boldsymbol{G}_{\boldsymbol{\theta}}^m(\tilde{\mathbf{z}}^{\boldsymbol{\Phi}}_{t},t,\mathbf{y})-\boldsymbol{\Phi}(\tilde{\mathbf{z}}_{t}^{\boldsymbol{\Phi}},t, s_m,\mathbf{y})||\\
  %         &=O((\Delta t)^p)(s_{m+1}-s_m),
  %     \end{aligned}
  % \end{equation}
\begin{equation}
    \sup_{t,s\in[s_m,s_{m+1})}||\mathbf{z}_0 - \mathbf{z}^{data}||\\
          =\mathcal{O}(\Delta t)(s_{m+1}-s_m).
  \end{equation}
  % $\mathbf{y}$ can be either unconditional or conditional.   
  \end{theorem}
  \begin{proof}
  The full proof, inspired by consistency model-related works \cite{lcm,pcm,consistent3d,connectingconsistency}, is provided in App. \ref{app:proof4th1}.
  \end{proof}
  \cref{th:1} shows that our SCTD loss achieves an arbitrarily small distillation error when the ODE solver’s step size $\Delta_t$ and the segment length $s_{m+1} - s_m$ are sufficiently small. By choosing an appropriate segment number $N_s$, we obtain a tighter error upper bound, which improves upon the bound $\mathcal{O}(\Delta t)(T)$ in CDS \cite{consistent3d} and $\mathcal{O}(\Delta t)(T-e)$ in GCS \cite{connectingconsistency}. This refinement enables more accurate gradients for optimizing the 3D representation $\boldsymbol{\theta}$, leading to higher-quality generative results. 
\section{Experiments}
\begin{figure*}[!t]
    \centering
     \includegraphics[width=\linewidth]{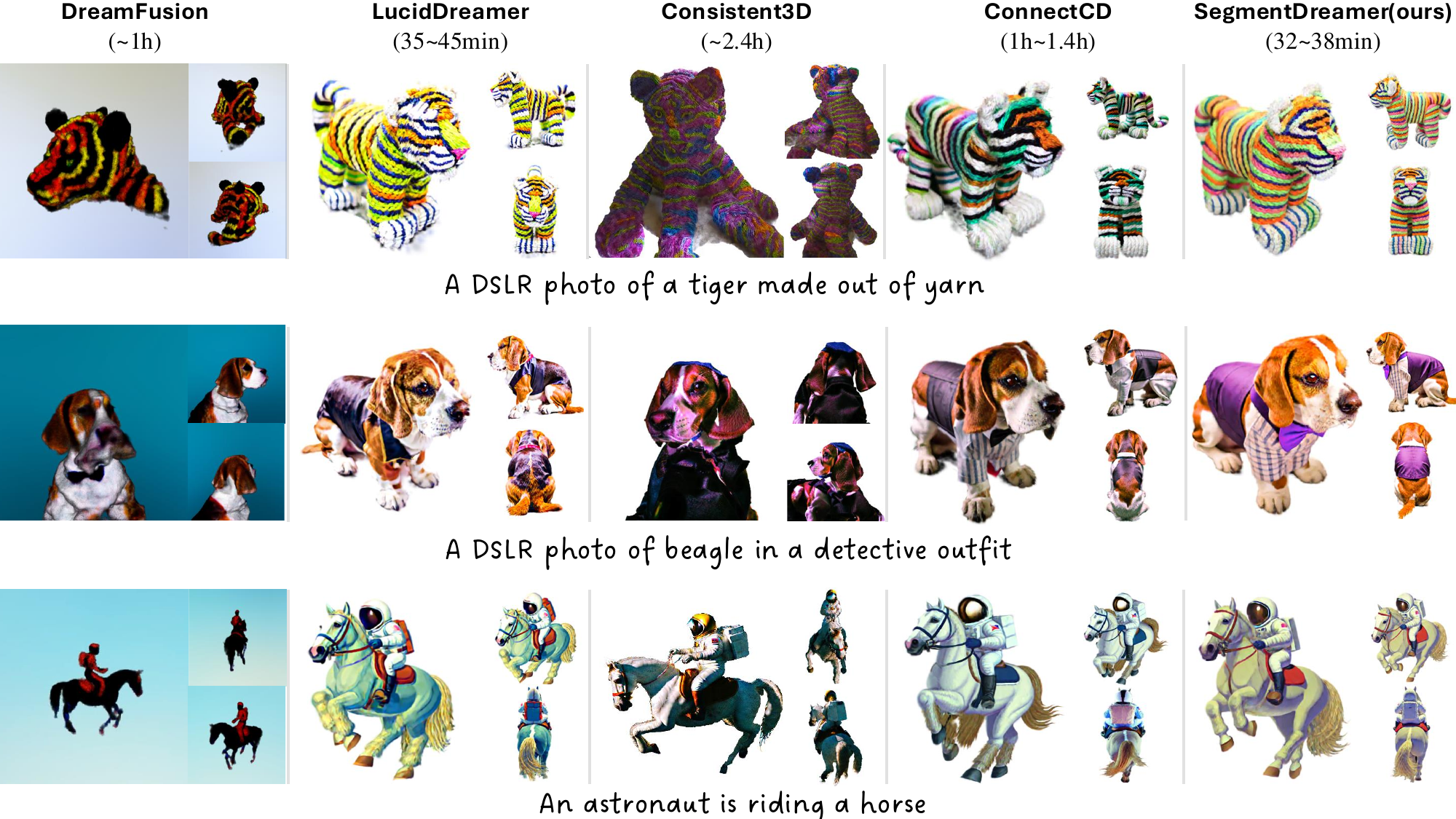}
     \caption{Qualitative comparison results of two SD-based metheds: DreamFusion \cite{dreamfusion} and LucidDreamer \cite{luciddreamer}, and two CD-based methods: Consistent3D \cite{consistent3d} and ConnectCD \cite{connectingconsistency}. CFG scales are set to 100, 7.5, 20$\sim$40, 7.5, 7.5, respectively. Our approach yields high-quality results within the shortest time. Please zoom in for details.}
     \label{fig:sctd_compare}
  \end{figure*}
\subsection{Implementation Details}
Our algorithm is implemented in PyTorch and trained on an A100 GPU for 5,000 iterations using the Adam optimizer \cite{adam}. We utilize Stable Diffusion 2.1 \cite{stablediffusion} as our foundational generative model and employ Point-E \cite{pointe} to initialize 3D Gaussians. Hyperparameters related to 3D Gaussian Splatting (3DGS), such as learning rates and camera poses, are set following the guidelines in previous works \cite{luciddreamer,connectingconsistency}. We have $t \sim \mathcal{U}(20,500+t_{warm})$, where $t_{warm}$ decreases linearly from 480 to 0 during the first 1,500 epochs.
\subsection{Text-to-3D Generation}
\noindent\textbf{Qualitative Comparison.}
We present a collection of 3D assets generated by SegmentDreamer from different camera views in \cref{fig:fisrtvis}. Our algorithm effectively produces high-quality, text-aligned 3D assets. To highlight the superior generative capability of SegmentDreamer, we conduct qualitative experiments comparing it with state-of-the-art methods based on Consistent Distillation (CD)—namely, Consistent3D \cite{consistent3d} and Connect3D \cite{connectingconsistency}—as well as Score Distillation (SD)-based methods, including DreamFusion \cite{dreamfusion} and LucidDreamer \cite{luciddreamer}, as illustrated in \cref{fig:sctd_compare}. For a fair comparison, all methods use Stable Diffusion 2.1 as the foundational generative model. As shown in \cref{fig:sctd_compare}, both DreamFusion and Consistent3D suffer from severe Janus artifacts and fail to generate accurate conditional outputs. LucidDreamer encounters issues with overexposure and ineffective conditional generation. Connect3D performs well in both conditional generation and color rendering; however, its results still contain undesired artifacts, particularly in the tiger and beagle models (see \cref{fig:sctd_compare}). Moreover, as a 3DGS-based method, it requires over an hour to generate a single 3D asset, limiting its practical usability. In contrast, SegmentDreamer can generate high-quality results within about 32 minutes using CFG \cite{cfg} and 38 minutes using Perp-Neg \cite{perp-neg}, proving its efficiency and superiority. More comparison results are shown in our supplementary material.

\begin{figure}[!t]
  \centering
   \includegraphics[width=\linewidth]{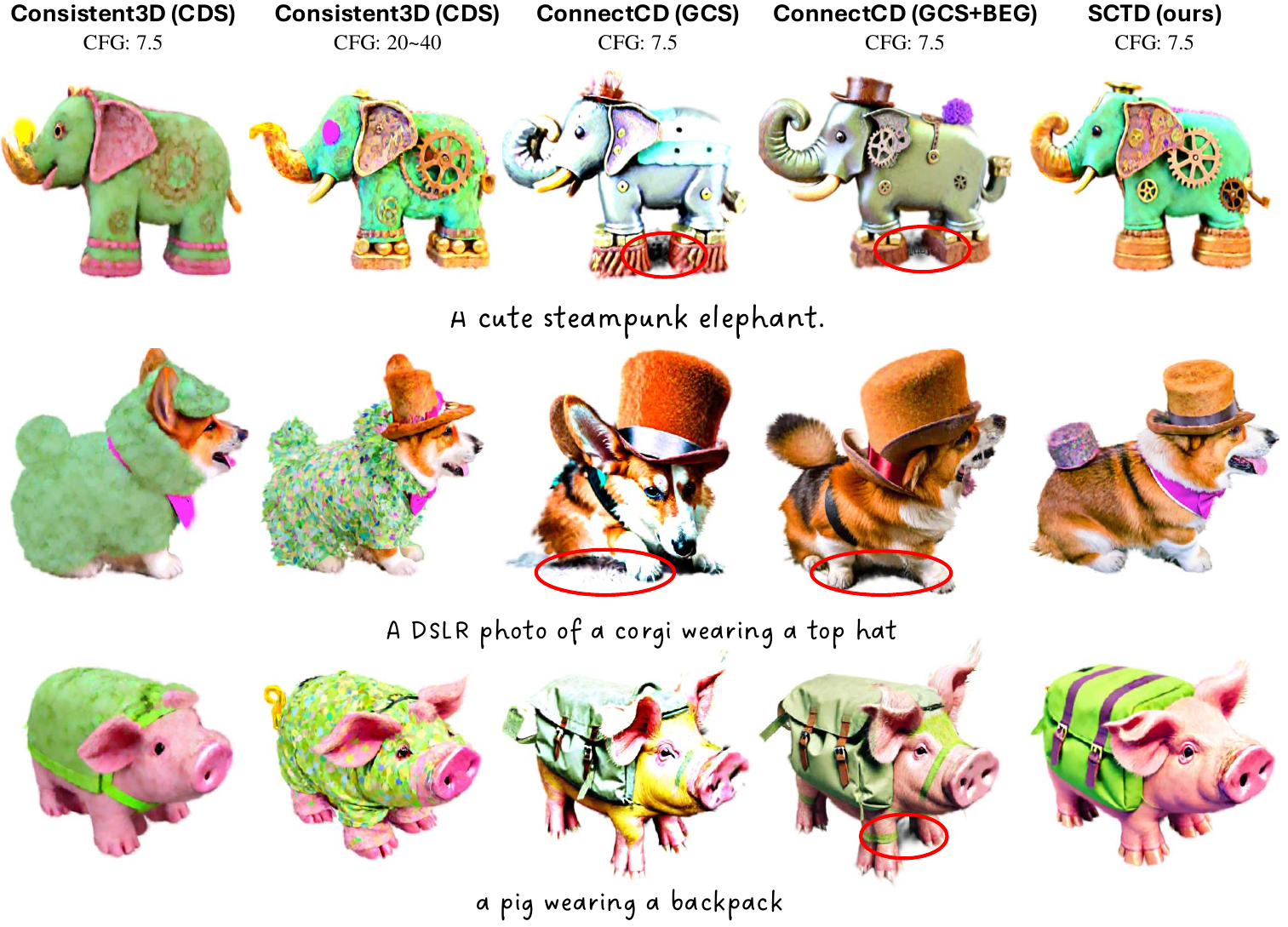}
   \caption{Visual comparison results of CDS \cite{consistent3d}, GCS \cite{connectingconsistency}, and GCS+BEG \cite{connectingconsistency}. Artifacts are marked with red circles.}
   \label{fig:loss_compare}
\end{figure}

We also compare our SCTD loss with existing CD-based losses, namely CDS \cite{consistent3d} and GCS \cite{connectingconsistency}. For a fair comparison, we use Stable Diffusion 2.1 as the generative model and keep the random seed, guidance scale, and 3DGS-related parameters consistent across all methods. As shown in \cref{fig:loss_compare}, CDS struggles to provide effective conditional guidance at a normal CFG scale, as also noted in \cite{connectingconsistency}. Even at a higher CFG scale, its results remain unsatisfactory due to the lack of cross-consistency in its distillation loss. GCS, meanwhile, suffers from overexposure and artifacts due to excessive conditional guidance. While it attempts to address this with Brightness-Equalized Generation (BEG), artifacts persist (see \cref{fig:loss_compare}; zoom in for details). In contrast, our SCTD mitigates ineffective conditional generation, overexposure, and artifacts, striking a better balance between self-consistency and cross-consistency.
\begin{figure}[!t]
  \centering
   \includegraphics[width=\linewidth]{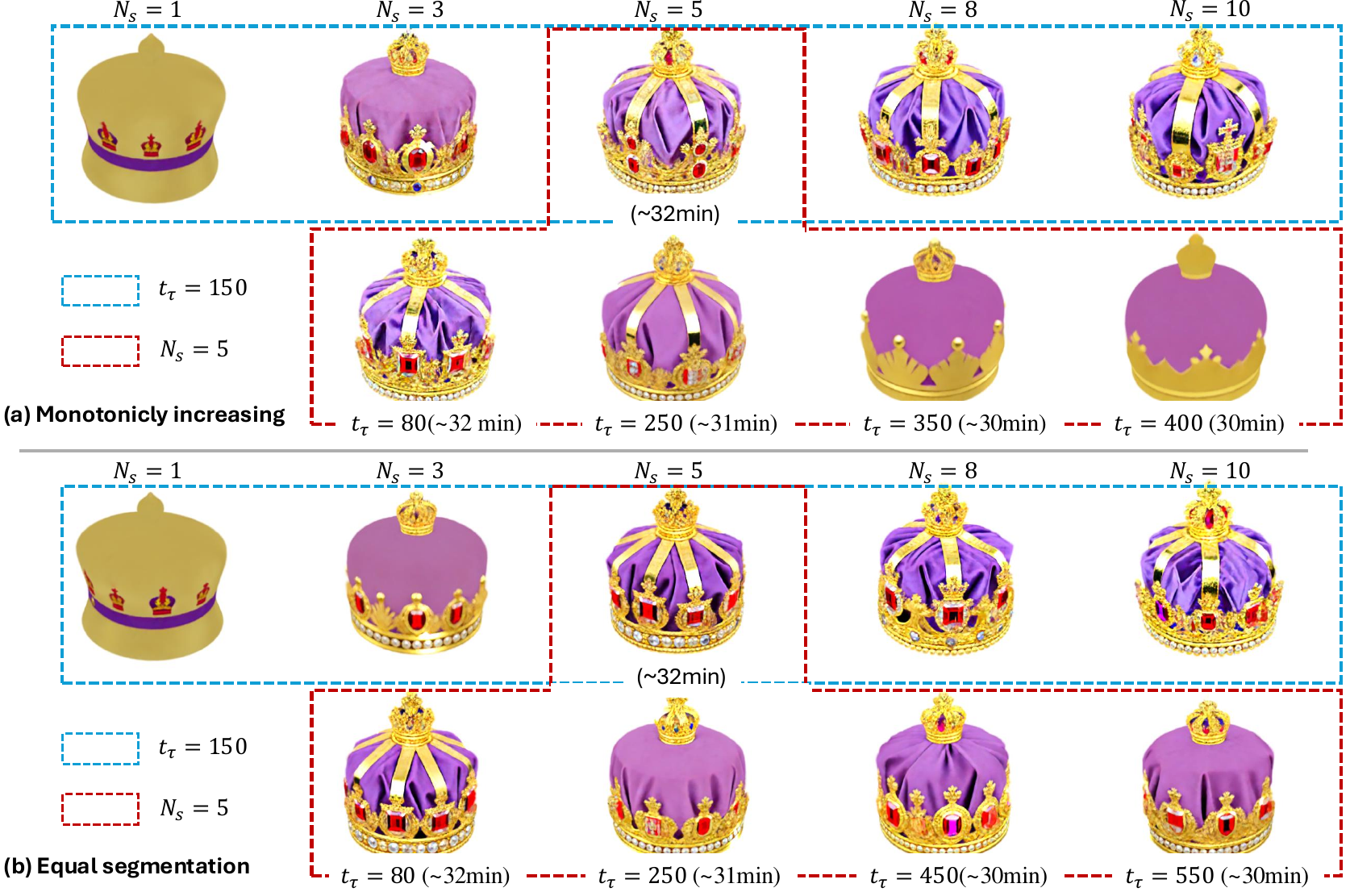}
   \caption{Ablation study of the trajectory segmentation strategy with different segment numer $N_s$ and time theshold $t_\tau$. Prompt: \textit{a DSLR photo of the Imperial State Crown of England.}}
   \label{fig:ablation_study_m}
\end{figure}
\begin{figure}[!t]
  \centering
   \includegraphics[width=\linewidth]{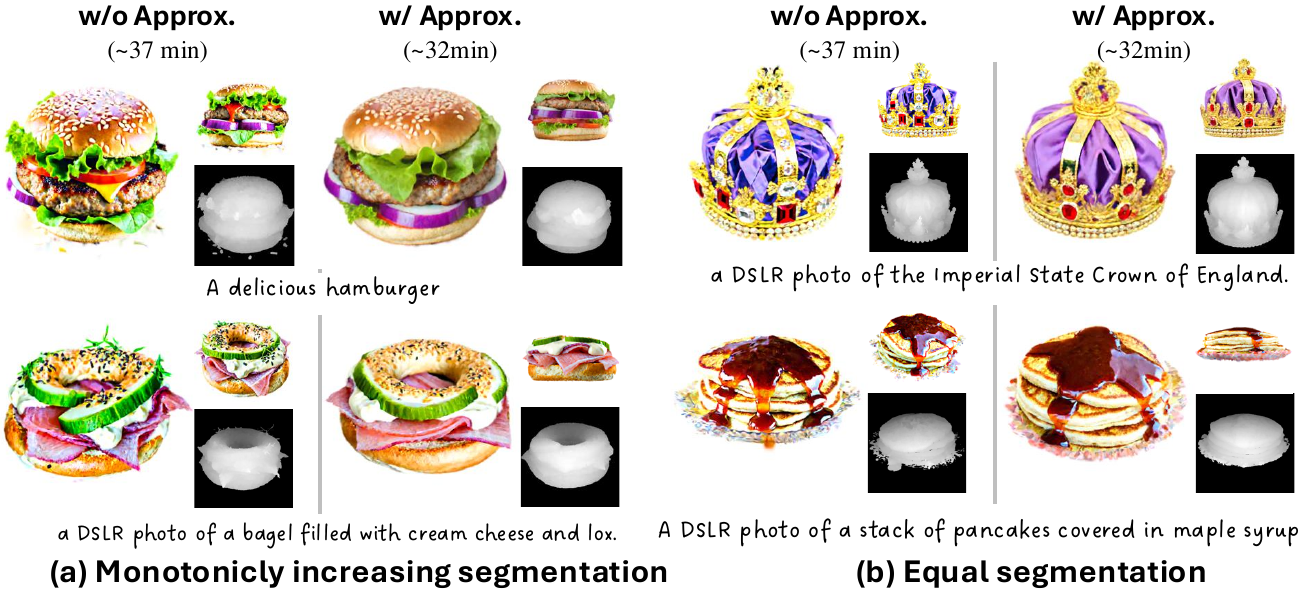}
   \caption{Visual comparison results before and after performing the approximation strategy proposed in \cref{secsec:tscd_loss}. ``\emph{w/ Approx.}'' and ``\emph{w/o Approx.}'' indicate whether the approximation strategy is used or not. The GFG scale is set to 7.5 for all generations.}
   \label{fig:ablation_app}
\end{figure}

\noindent\textbf{Quantitative Comparison.}
We evaluate the text alignment and human preference of each text-to-3D method as follows: we generate 3D models from 40 prompts selected from the DreamFusion Gallery \footnote{https://dreamfusion3d.github.io/gallery.html}, and calculate the average CLIP Score and ImageReward (IR) score across 120 camera views for each model. Following Connect3D \cite{connectingconsistency}, we compute the FID score between 40 objects (with 120 rendered views for each) and 50,000 images generated by Stable Diffusion 2.1. As shown in the results, our method achieves the best performance in text alignment, human preference, and fidelity, while requiring the shortest generation time, demonstrating its superior overall performance. The 40 prompts used are provided in the supplementary material.
% We assess text alignment and human preference for each text-to-3D method in \cref{tab:sctd_compare}, where we generate 3D models from 40 prompts selected from the DreamFusion Gallery \footnote{https://dreamfusion3d.github.io/gallery.html} and compute the average CLIP Score and ImageReward score across 120 camera views per model.
% As shown, our method achieves the best text alignment and human preference while requiring the least time, demonstrating its superior performance. The 40 prompts are given in the supplementary material.

\noindent\textbf{User Study.}
We conduct a user study involving 40 volunteers to evaluate the generation quality of each algorithm. Each volunteer is asked to review eight groups of five 360-degree rendered videos and to rank them based on three criteria: text alignment (\textbf{Q1}), object realism (\textbf{Q2}), and object details (\textbf{Q3}; e.g., artifacts, brightness, and the Janus problem). To minimize bias, the order of the videos is randomized. As shown in \cref{tab:sctd_compare}, users consistently prefer our method across all criteria.
% We conduct a user study with 40 volunteers to assess human preference for each algorithm’s results. Each volunteer reviews eight groups of five 360-degree rendered videos, ranking them from the perspective of prompt alignment, fidelity, and detail. To minimize bias, video order is randomized. As shown in \cref{tab:sctd_compare}, users consistently favor our method across all criteria.
\section{Ablation Study}
\begin{table}[!t]
\caption{
Quantitative comparison using CLIP score, IR score, training time (min), FID score, and three user study questions given by 40 volunteers. \textbf{Bold} and \underline{underline} texts denote the best and second-best performance, respectively.}
\centering
\setlength{\tabcolsep}{1.1mm}
\scriptsize
\begin{tabular}{lccccccc}
\toprule
                                        &CLIP-L$\uparrow$         &  IR $\uparrow$       & FID $\downarrow$                & Time $\downarrow$       & \textbf{Q1}$\downarrow$   & \textbf{Q2}$\downarrow$ & \textbf{Q3}$\downarrow$\\\midrule
DreamFusion\cite{dreamfusion}           &  28.47                  &  -0.004              &  140.84                         & 60                      & 4.73                      & 4.87                    & 4.90           \\ 
LucidDreamer\cite{luciddreamer}         &  29.99                  &  0.006               &  121.80                         & \underline{45}          & 2.93                      & 2.98                    & 2.93           \\
Consistent3D\cite{consistent3d}         &  30.60                  &  0.004               &  113.61                         & 140                     & 4.14                      & 3.88                    & 3.93           \\ 
ConnectCD \cite{connectingconsistency}  &  \underline{30.73}      &  \underline{0.018}   &  \underline{112.61}             & 80                      & \underline{1.63}          & \underline{1.80}        & \underline{1.93}           \\
\rowcolor{MyGray}
\textbf{SegmentDreamer}                 &\textbf{30.88}           &  \textbf{0.020}      &  \textbf{110.45}                & \textbf{38}             & \textbf{1.57}             & \textbf{1.47}           & \textbf{1.33}           \\
\bottomrule
      \end{tabular}
  \label{tab:sctd_compare}
\end{table}
\label{sec:ablationstudy}
\noindent\textbf{Effect of Trajectory Segmentation Strategy}
We visualize the effect of different trajectory segmentation strategies applied to our algorithm in \cref{fig:ablation_study_m}. When the segment number $N_s$ and the time threshold $t_\tau$ are fixed, we observe that the segmentation strategy has minimal impact on the generated results. This is primarily because the segmentation strategy essentially determines $\sup|t-e|$. However, since $t$ and $s$ are randomly selected, $|t-s_m|$, $|t-s|$, and $|s-s_m|$ vary randomly during each training iteration, and their values can be either large or small.
 
\noindent\textbf{Effect of Segment Number $N_s$}. We visualize the effect of different segment numbers $N_s$ in \cref{fig:ablation_study_m}. As observed, a small $N_s$, regardless of the segmentation strategy, consistently results in overly smoothed outputs. As $N_s$ increases, the generation quality improves. However, when $N_s$ becomes too large (\emph{e.g.} 10), while more details emerge, the overall model representation becomes vague—a phenomenon also noted in LucidDreamer  \cite{luciddreamer}. We find that $N_s = 5$ performs well in most cases.

\noindent\textbf{Effect of time threshold $t_\tau$}. As shown in \cref{fig:ablation_study_m}, a large $t_\tau$ leads to overly smoothed results for both segmentation strategies. This is because a limited number of sampling steps in the diffusion process fail to effectively preserve the information of the initial point $\mathbf{z}_0$, highlighting the necessity of incorporating multi-step deterministic sampling.

\noindent\textbf{Validation of Approximation Strategy}. We analyze the impact of replacing $\boldsymbol{G}_{\boldsymbol{\theta}}(\tilde{\mathbf{z}}_t^{\boldsymbol{\Phi}},t,\mathbf{y})$ with $\mathbf{z}_{s_m}$ in \cref{fig:ablation_app}. As shown, $\boldsymbol{G}_{\boldsymbol{\theta}}(\tilde{\mathbf{z}}_t^{\boldsymbol{\Phi}},t,\mathbf{y})$ introduces slight overexposure and geometric distortion. In contrast, $\mathbf{z}_{s_m}$ not only produces better generation results but also reduces the training time. This suggests a gap between theory and practice, similar to the omission of U-Net Jacobian in DreamFusion \cite{dreamfusion}.
\section{Conclusion}
% In this paper, we analyze the improper conditional guidance in previous Consistency Distillation (CD)-based methods, identifying the imbalance between self-consistency and cross-consistency as the root cause. We also show that large distillation error upper bounds limit generative capabilities. To address this, we propose the Segmented Consistency Trajectory Distillation loss, which improves conditional guidance and tightens the distillation upper bound. Experimental results demonstrate that our SegmentDreamer achieves superior generation quality within a reasonable time, advancing distillation-based text-to-3D methods toward practical use.
We analyze improper conditional guidance in previous Consistency Distillation methods and identify its root cause as an imbalance between self-consistency and cross-consistency in denoising trajectories. Additionally, we find that large distillation error upper bounds limit generative capability. To address this, we propose the Segmented Consistency Trajectory Distillation loss, which improves conditional guidance and tightens the distillation bound. Experiments show that our SegmentDreamer achieves high-quality generation efficiently, advancing distillation-based text-to-3D methods toward practical use.

\noindent\textbf{Limitation.} SegmentDreamer mainly focuses on single-instance generation and performs suboptimally in multi-instance generation.

\noindent\textbf{Potential Negative Impact.} SegmentDreamer focuses on content synthesis, which may be used to create misleading content.
\section*{Acknowledgements}
% \small
This project is supported by the National Natural Science Foundation of China (12326618, U22A2095), the National Key Research and Development Program of China (2024YFA1011900), the Major Key Project of PCL (PCL2024A06), and the Project of Guangdong Provincial Key Laboratory of Information Security Technology (2023B1212060026).

{
    \small
    \bibliographystyle{ieeenat_fullname}
    \bibliography{main}
}

\clearpage
\setcounter{page}{1}
\maketitlesupplementary
\setcounter{section}{0} % 重置章节计数器
\setcounter{figure}{0}
\setcounter{table}{0}
\setcounter{equation}{0}
\renewcommand{\thesection}{\Alph{section}}
To provide proofs of SegmentDreamer and visual comparisons of state-of-the-arts, this supplementary material includes the following contents:
\begin{itemize}[leftmargin=15pt]
    \item \Cref{app:gcs}: Guided Consistency Sampling Loss
    \item \Cref{app:flaw}: Flaws in the Consistency Function of Guided Consistency Sampling
    \item \Cref{app:oversaturation}: Oversaturation and Artifacts in
GCS
    \item \Cref{app:proof4SCTD}: How to Connect SCTD with SDS
    \item \Cref{app:proof4crt}: Computation Reduction Trick
    \item \Cref{app:proof4th1}: Proof of Upper Bound of Distillation Error
    \item \Cref{app:visual}: Visual Comparisons of State-of-the-arts
    % \item \Cref{app:limit}: Limitation and Potential Negative Impacts
\end{itemize}

\section{Guided Consistency Sampling Loss}
\label{app:gcs}
The Guided Consistency Sampling (GCS) loss \cite{connectingconsistency} is composed of a compact consistency loss $\mathcal{L}_{\text{CC}}$, a conditional guidance loss $\mathcal{L}_{\text{CG}}$, and a pixel-wise constraint loss $\mathcal{L}_{\text{CP}}$, which are defined by
\begin{equation}
  \label{eq:gcsloss}
\begin{aligned}
&\mathcal{L}_{\text{CC}}(\boldsymbol{\theta})= \mathbb{E}[\left\| \boldsymbol{G_\theta}(\tilde{\mathbf{z}}^{\boldsymbol{\Phi}}_{t}, t, e, \emptyset) - \boldsymbol{G_\theta}(\hat{\mathbf{z}}_s^{\boldsymbol{\Phi}},s, e, \emptyset) \right\|_2^2],\\
&\mathcal{L}_{\text{CG}}(\boldsymbol{\theta})= \mathbb{E}[\left\| \boldsymbol{F_\theta}(\mathbf{z}_{e}, e, \emptyset) - \boldsymbol{F_\theta}(\boldsymbol{G}_\theta(\tilde{\mathbf{z}}^{\boldsymbol{\Phi}}_t, t,e,\mathbf{y}),e, \emptyset) \right\|_2^2],\\
&\mathcal{L}_{\text{CP}}(\boldsymbol{\theta})\!= \!\mathbb{E}[\left\|\! \boldsymbol{D}(\!F_\theta(\mathbf{z}_{e}, e, \emptyset)) \!\!- \!\!\boldsymbol{D}(\boldsymbol{F_\theta}(\boldsymbol{G}_\theta(\tilde{\mathbf{z}}^{\boldsymbol{\Phi}}_t, t,e,\mathbf{y}),e,\mathbf{y})\!)\! \right\|_2^2],\\
\end{aligned}
\end{equation}
where $e<s<t$, $\tilde{\mathbf{z}}^{\boldsymbol{\Phi}}_{t}$ is estimated by the following trajectory: $\mathbf{z}_{e}=\alpha_t\mathbf{z}_{0}+\sigma_t\boldsymbol{\epsilon}^*\to\tilde{\mathbf{z}}^{\boldsymbol{\Phi}}_{s}=\boldsymbol{\Phi}(\mathbf{z}_{e},e, s,\emptyset)\to\tilde{\mathbf{z}}^{\boldsymbol{\Phi}}_{t}=\boldsymbol{\Phi}(\tilde{\mathbf{z}}^{\boldsymbol{\Phi}}_{s},s, t,\emptyset)$, $\hat{\mathbf{z}}^{\boldsymbol{\Phi}}_{s}=\boldsymbol{\Phi}(\tilde{\mathbf{z}}^{\boldsymbol{\Phi}}_{t}, t, s, \mathbf{y})$, and $\boldsymbol{D}$ denotes the VAE decoder.
\section{Flaws in the Consistency Function of Guided Consistency Sampling}
\label{app:flaw}
As we know, given a well-trained diffusion model $\boldsymbol{\phi}$, there exists an exact solution from timestep $t$ to $e$ \cite{dpm}: 
\begin{equation}
    \label{eq:dpm-solver-o}
    \boldsymbol{G}(\mathbf{z}_t,t,e,\mathbf{y})\!=\!\frac{\alpha_e}{\alpha_t}\mathbf{z}_t+\alpha_s\!\int_{\lambda_t}^{\lambda_e}e^{-\lambda}\boldsymbol{\epsilon_\phi}(\mathbf{z}_{t_\lambda(\lambda)},t_\lambda(\lambda),\mathbf{y})\mathrm{d}\lambda,
\end{equation}
where $\lambda_t=\mathrm{ln}\frac{\alpha_t}{\sigma_t}$ and $t_\lambda$ denotes the inverse function of $\lambda_t$. Inspired by \cite{pcm}, we find \cref{eq:ccloss} suggests that GCS aims to optimize a 3D representation $\boldsymbol{\theta}$ such that $\boldsymbol{G_\theta}(\mathbf{z}_t, t, e, \emptyset) = \boldsymbol{G_\theta}(\hat{\mathbf{z}}_s^{\boldsymbol{\Phi}},s, e, \emptyset) $ for $\forall t,s,e\in [0,T]$ where $t>s>e$. This implies $\boldsymbol{\epsilon}_{\boldsymbol{\phi}}(\mathbf{z}_t,t,\mathbf{y})$, as defined in \cref{eq:gcs_cf}, is not an approximation but an exact solution learning related to the 3D representation $\boldsymbol{\theta}$, \emph{i.e.}, $\boldsymbol{\epsilon}_{\boldsymbol{\phi}}(\mathbf{z}_t,t,\mathbf{y})=\frac{\int_{\lambda_t}^{\lambda_e}e^{-\lambda}\boldsymbol{\epsilon_\phi}(\mathbf{z}_{t_\lambda(\lambda)}^{\mathbf{c}},t_\lambda(\lambda),\mathbf{y})\mathrm{d}\lambda}{\int_{\lambda_t}^{\lambda_e}e^{-\lambda}\mathrm{d}\lambda}$. However, dropping the target timestep $e$ in $\boldsymbol{\epsilon}_{\boldsymbol{\phi}}(\mathbf{z}_t,t,\mathbf{y})$, as GCS does, is problematic. Suppose we predict both $\mathbf{z}_{e}$ and $\mathbf{z}_{e'}$ from $\mathbf{z}_t$ where $t>e'>e$, we must have
\begin{equation}
  \label{eq:gcs_flaw}
  \begin{aligned}
    &   \boldsymbol{\epsilon}_{\boldsymbol{\phi}}(\mathbf{z}_t,t,\mathbf{y})=\frac{\int_{\lambda_t}^{\lambda_{e'}}e^{-\lambda}\boldsymbol{\epsilon_\phi}(\mathbf{z}_{t_\lambda(\lambda)},t_\lambda(\lambda),\mathbf{y})\mathrm{d}\lambda}{\int_{\lambda_t}^{\lambda_{e'}}e^{-\lambda}\mathrm{d}\lambda}\\
    &  \boldsymbol{\epsilon}_{\boldsymbol{\phi}}(\mathbf{z}_t,t,\mathbf{y})=\frac{\int_{\lambda_t}^{\lambda_e}e^{-\lambda}\boldsymbol{\epsilon_\phi}(\mathbf{z}_{t_\lambda(\lambda)},t_\lambda(\lambda),\mathbf{y})\mathrm{d}\lambda}{\int_{\lambda_t}^{\lambda_e}e^{-\lambda}\mathrm{d}\lambda}.\\
  \end{aligned}
\end{equation} 
Clearly, without the target timestep in $\boldsymbol{\epsilon}_{\boldsymbol{\phi}}(\mathbf{z}_t,t,\mathbf{y})$, optimizing a 3D representation $\boldsymbol{\theta}$ to satisfy both conditions in \cref{eq:gcs_flaw} for all intervals $[e, t]$ is invalid. For GCS, as the number of training steps increases, the above unreasonable phenomena occur frequently, potentially resulting in poor distillation results.
\section{Oversaturation and Artifacts in GCS}
\begin{figure}[!t]
\setlength{\abovecaptionskip}{-0.15mm}
    \centering
   %\includegraphics[width=8.6cm,height=4.2cm]{fig/framework.pdf}
   \includegraphics[width=\linewidth]{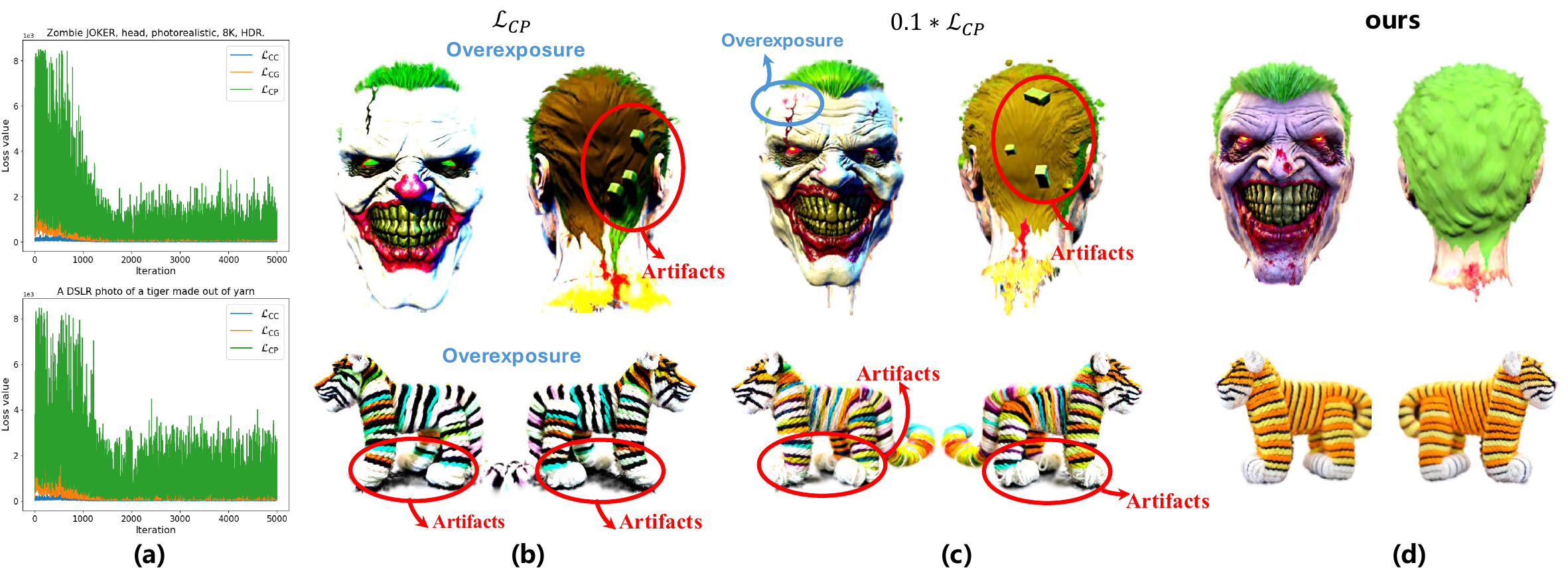}
   \caption{Analysis on self- and cross-consistency in GCS (a), and visual analysis of oversaturation and artifacts (b)-(d).}
    \label{app:oversat}
    \vspace{-1em}
\end{figure}
\label{app:oversaturation}
\cref{app:oversat} shows that the cross-consistency loss $\mathcal{L}_{\text{CG}}+\mathcal{L}_{\text{CP}}$ (with $\mathcal{L}_{\text{CP}}$ dominating) greatly exceeds the self-consistency loss $\mathcal{L}_{\text{CC}}$. As shown in \cref{app:oversat}, we scale down $\mathcal{L}_{\text{CP}}$ \textbf{(c)} and observe that the oversaturation \textbf{(b)} is alleviated, but undesirable geometry persists (red circle). This suggests that such oversaturation is brought from an excessively large value of cross-consistency loss, \textit{i.e.,} the ``excessive conditional guidance,'' which has been analyzed in Sec. \textcolor{iccvblue}{4.1}.
In contrast, our $\mathcal{L}_\text{SCTD}$ \textbf{(d)} successfully addresses the above issues.
\section{How to Connect SCTD with SDS?}
\label{app:proof4SCTD}
As described in \cref{secsec:tscd_loss}, given any subtrajectory $[s_m,s_{m+1}]$, $\boldsymbol{G}_{\boldsymbol{\theta}}(\mathbf{z}_{t},t,s_m,\mathbf{y}):=\boldsymbol{G}_{\boldsymbol{\theta}}^m(\mathbf{z}_{t},t,\mathbf{y})=\frac{\alpha_{s_m}}{\alpha_{t}}\mathbf{z}_{t}-\alpha_{s_m}\boldsymbol{\epsilon}_{\boldsymbol{\phi}}(\mathbf{z}_{t},t,\mathbf{y})\int_{\lambda_{t}}^{\lambda_{s_m}}e^{-\lambda}\mathrm{d}\lambda$. Then, we have
\begin{equation}
    \label{eq:app1}
    \boldsymbol{\epsilon}_{\boldsymbol{\phi}}(\mathbf{z}_{t},t,\mathbf{y})=\frac{\boldsymbol{G}_{\boldsymbol{\theta}}^m(\mathbf{z}_{t},t,\mathbf{y})-\frac{\alpha_{s_m}}{\alpha_{t}}\mathbf{z}_{t}}{\alpha_{s_m}\int_{\lambda_{t}}^{\lambda_{s_m}}e^{-\lambda}\mathrm{d}\lambda},
\end{equation}
where $\mathbf{z}_t=\alpha_t\mathbf{z}_0+\sigma_t\boldsymbol{\epsilon}$. In this case, $\mathbf{y}$ can represent any prompt embedding, including $\emptyset$. According to \cref{eq:sds-cfg}, $\mathcal{L}_{\text{SDS}}$ can be further transformed into:
\begin{equation}
  \label{eq:sds-cs}
  % \begin{aligned}
  %   &\hat{\epsilon}(\mathbf{z}_t;y,t)-\epsilon = \epsilon_{\phi}(\mathbf{z}_t;\emptyset,t)-\epsilon-\frac{\alpha_t \sigma_s}{\sigma_t \alpha_s}\epsilon_{\phi}(\mathbf{z}_s;\emptyset,t) \\
  %   &+\frac{\alpha_t \sigma_s}{\sigma_t \alpha_s}\epsilon_{\phi}(\mathbf{z}_s;\emptyset,t) +(\omega+1)(\epsilon_{\phi}(\mathbf{z}_t;y,t)-\epsilon_{\phi}(\mathbf{z}_t;\emptyset,t)),\\
  % \end{aligned}
  \begin{aligned}
    &\hat{\boldsymbol{\epsilon}}_{\boldsymbol{\phi}}(\mathbf{z}_{t},t,\mathbf{y})-\boldsymbol{\epsilon} = \boldsymbol{\epsilon_\phi}(\mathbf{z}_{t},t,\mathbf{y})-\boldsymbol{\epsilon}- \boldsymbol{G}^m_{\boldsymbol{\theta}}(\hat{\mathbf{z}}_{s}^{\boldsymbol{\Phi}},s,\mathbf{y}) +  \\
    & \boldsymbol{G}^m_{\boldsymbol{\theta}}(\hat{\mathbf{z}}_{s}^{\boldsymbol{\Phi}},s,\mathbf{y}) +\omega(\boldsymbol{\epsilon_\phi}(\mathbf{z}_{t},t,\mathbf{y})-\boldsymbol{\epsilon_\phi}(\mathbf{z}_{t},t,\emptyset)),\\
  \end{aligned}
\end{equation}
where $\hat{\mathbf{z}}_{s}^{\boldsymbol{\Phi}}=\boldsymbol{\Phi}(\mathbf{z}_t,t,s,\mathbf{y})$.
By substituting \cref{eq:app1} into \cref{eq:sds-cs}, we obtain
\begin{equation}
    \label{eq:app2}
      \begin{aligned}
        & \mathcal{L}_{\text{SDS}}(\boldsymbol{\theta})\!=\!\mathbb{E}_{t}[b(t)\!||\boldsymbol{G}^m_{\boldsymbol{\theta}}(\hat{\mathbf{z}}_{s}^{\boldsymbol{\Phi}},s,\mathbf{y})-\boldsymbol{G}^m_{\boldsymbol{\theta}}(\mathbf{z}_{t},t,\mathbf{y})\\
        &+\omega(\boldsymbol{G}^m_{\boldsymbol{\theta}}(\mathbf{z}_{t},t, \emptyset)\!-\!\boldsymbol{G}^m_{\boldsymbol{\theta}}(\mathbf{z}_{t},t,\mathbf{y}))\!-\!\boldsymbol{G}^m_{\boldsymbol{\theta}}(\hat{\mathbf{z}}_{s}^{\boldsymbol{\Phi}},s,\mathbf{y})\\
        &+\frac{\alpha_{s_m}}{\alpha_{t}}\mathbf{z}_{t}-\alpha_{s_m}\int_{\lambda_{t}}^{\lambda_{s_m}}e^{-\lambda}\mathrm{d}\lambda \boldsymbol{\epsilon}||_2^2],
      \end{aligned}
    \end{equation}
where $b(t)=\frac{\omega(t)}{(\alpha_{s_m}\int_{\lambda_{t}}^{\lambda_{s_m}}e^{-\lambda}\mathrm{d}\lambda)^2}$. Furthermore,
\begin{equation}
    \label{eq:app3}
    \begin{aligned}
        &\frac{\alpha_{s_m}}{\alpha_{t}}\mathbf{z}_{t}-\alpha_{s_m}\int_{\lambda_{t}}^{\lambda_{s_m}}e^{-\lambda}\mathrm{d}\lambda \boldsymbol{\epsilon}\\
        &=\frac{\alpha_{s_m}}{\alpha_{t}}(\alpha_{t}\mathbf{z}_0^{\mathbf{c}}+\sigma_{t}\boldsymbol{\epsilon})+\alpha_{s_m}(e^{\lambda_{s_m}-\lambda_{t}}-1)\boldsymbol{\epsilon}\\
        &=\alpha_{s_m}\mathbf{z}_0+\sigma_{s_m}\boldsymbol{\epsilon}\\
        &=\mathbf{z}_{s_m}
    \end{aligned}
\end{equation}
Substituting \cref{eq:app3} into \cref{eq:app2}, we can obtain \cref{eq:noise-cd-tscd}. Proof is completed.
\section{Computation Reduction Trick}
\label{app:proof4crt}
\cref{eq:sds-cfg} can be transformed into
\begin{equation}
    \label{eq:sds-cfg-mod}
    \begin{aligned}
        &\hat{\boldsymbol{\epsilon}}_{\boldsymbol{\phi}}(\mathbf{z}_t,t,\mathbf{y})\!-\!\boldsymbol{\epsilon}\!=\!\boldsymbol{\epsilon_\phi}(\mathbf{z}_t,t,\mathbf{y})\!-\!\boldsymbol{\epsilon}\!+\!\omega(\boldsymbol{\epsilon_\phi}(\mathbf{z}_t,t,\mathbf{y})\!-\!\boldsymbol{\epsilon_\phi}(\mathbf{z}_t,t,\emptyset))\\
        & 
        =\!\boldsymbol{\epsilon_\phi}(\mathbf{z}_t,t,\emptyset)\!-\!\boldsymbol{\epsilon}\!+\!(\omega+1)(\boldsymbol{\epsilon_\phi}(\mathbf{z}_t,t,\mathbf{y})\!-\!\boldsymbol{\epsilon_\phi}(\mathbf{z}_t,t,\emptyset))\\ 
    \end{aligned}
\end{equation}
Based on \cref{eq:sds-cfg-mod}, we can readily derive \cref{eq:noise-cd-tscd-mod} by following the procedure described in App. \ref{app:proof4SCTD}.
\section{Proof of Upper Bound of Distillation Error}
\label{app:proof4th1}
Before proving \cref{th:1}, we first give the following lemma:
\begin{lemma}
    \label{th:upperbound}
    Given a sub-trajectory $[s_m,s_{m+1}]$, let $\Delta t=max_{t,s\in[s_m,s_{m+1})}\{|t-s|\}$. We assume $\boldsymbol{G}^m_{\boldsymbol{\theta}}$ satisfies the Lipschitz condition and the ODE solver has local error uniformally bounded by $O(t-s)^{p+1}$ with $p\geq 1$. If $\boldsymbol{G}^m_{\boldsymbol{\theta}}(\tilde{\mathbf{z}}_{t}^{\boldsymbol{\Phi}},t,\emptyset)=\boldsymbol{G}^m_{\boldsymbol{\theta}}(\hat{\mathbf{z}}_{s}^{\boldsymbol{\Phi}},s,\emptyset)$ for $\forall t, s\in[s_m,s_{m+1}]$, we have 
    \begin{equation}
        \begin{aligned}
            &\sup_{t,s\in[s_m,s_{m+1})}\!\!\!\!\!||\boldsymbol{G}_{\boldsymbol{\theta}}^m(\tilde{\mathbf{z}}^{\boldsymbol{\Phi}}_{t},t,\mathbf{y})-\boldsymbol{\Phi}(\tilde{\mathbf{z}}_{t}^{\boldsymbol{\Phi}},t, s_m,\mathbf{y})||\\
            &=O((\Delta t)^p)(s_{m+1}-s_m).
        \end{aligned}
    \end{equation}   
    \end{lemma}
    \begin{proof}
        % We first define $\boldsymbol{G}_{\boldsymbol{\theta}}^{m,m'}(z_t,t,\mathbf{y})=
        % \boldsymbol{G}_{\boldsymbol{\theta}}^{m'}(\cdots\boldsymbol{G}_{\boldsymbol{\theta}}^{m-1}(\boldsymbol{G}_{\boldsymbol{\theta}}^m(\mathbf{z}_t,s_m,\mathbf{y}),s_{m-1},\mathbf{y})\cdots,s_{m'},\mathbf{y})$ that transform any point $\mathbf{z}_t$ on $m$-th sub-trajectory to the solution of $m'$-th subtrajectory.
        The proof is based on \cite{cm,pcm,connectingconsistency,consistent3d}. Let 
        \begin{equation}
            \boldsymbol{e}_{n-1}:=\boldsymbol{G}_{\boldsymbol{\theta}}^m(\tilde{\mathbf{z}}_{s}^{\boldsymbol{\Phi}},s,\mathbf{y})-\boldsymbol{\Phi}(\tilde{\mathbf{z}}_{s}^{\boldsymbol{\Phi}},s, s_m,\mathbf{y}),
        \end{equation}
    where $\tilde{\mathbf{z}}_{s}^{\boldsymbol{\Phi}}=\boldsymbol{\Phi}(\mathbf{z}_{s_m},s_m, s,\mathbf{y})$. According to the condition, we have 
        \begin{equation}
            \begin{aligned}
                &\boldsymbol{e}_n=\boldsymbol{G}_{\boldsymbol{\theta}}^m(\tilde{\mathbf{z}}_{t}^{\boldsymbol{\Phi}},t,\mathbf{y})-\boldsymbol{\Phi}(\tilde{\mathbf{z}}_{t}^{\boldsymbol{\Phi}},t, s_m,\mathbf{y})\\
                &=\boldsymbol{G}_{\boldsymbol{\theta}}^m(\hat{\mathbf{z}}_{s}^{\boldsymbol{\Phi}},s,\mathbf{y})-\boldsymbol{G}_{\boldsymbol{\theta}}^m(\tilde{\mathbf{z}}_{s}^{\boldsymbol{\Phi}},s,\mathbf{y})\\
                &+\boldsymbol{G}_{\boldsymbol{\theta}}^m(\tilde{\mathbf{z}}_{s}^{\boldsymbol{\Phi}},s,\mathbf{y})-\boldsymbol{\Phi}(\tilde{\mathbf{z}}_{s}^{\boldsymbol{\Phi}},s, s_m,\mathbf{y})\\
                &=\boldsymbol{G}_{\boldsymbol{\theta}}^m(\hat{\mathbf{z}}_{s}^{\boldsymbol{\Phi}},s,\mathbf{y})-\boldsymbol{G}_{\boldsymbol{\theta}}^m(\tilde{\mathbf{z}}_{s}^{\boldsymbol{\Phi}},s,\mathbf{y})+\boldsymbol{e}_{n-1},\\
            \end{aligned}
        \end{equation}
    Provided that $\boldsymbol{G}_{\boldsymbol{\theta}}^m$ satisfies $L$-Lipschitz condition, we have
        \begin{equation}
            \begin{aligned}
                &||e_n||=||e_{n-1}+\boldsymbol{G}_{\boldsymbol{\theta}}^m(\hat{\mathbf{z}}_{s}^{\boldsymbol{\Phi}},s,\mathbf{y})-\boldsymbol{G}_{\boldsymbol{\theta}}^m(\tilde{\mathbf{z}}_{s}^{\boldsymbol{\Phi}},s,\mathbf{y})||\\
                &\leq ||e_{n-1}||+||\boldsymbol{G}_{\boldsymbol{\theta}}^m(\hat{\mathbf{z}}_{s}^{\boldsymbol{\Phi}},s,\mathbf{y})-\boldsymbol{G}_{\boldsymbol{\theta}}^m(\tilde{\mathbf{z}}_{s}^{\boldsymbol{\Phi}},s,\mathbf{y})||\\
                &\leq \boldsymbol{e}_{n-1}+L||\hat{\mathbf{z}}_{s}^{\boldsymbol{\Phi}}-\tilde{\mathbf{z}}_{s}^{\boldsymbol{\Phi}}||\\
                &\leq \boldsymbol{e}_{n-1}+L\cdot O((t-s)^{p+1})\\
                &\leq \boldsymbol{e}_{n-1}+L(t-s)\cdot O((\Delta t)^p).\\
            \end{aligned}
        \end{equation}
        Besides, according to the boundray condition,
        \begin{equation}
            \begin{aligned}
                &\boldsymbol{e}_{s_m}=\boldsymbol{G}_{\boldsymbol{\theta}}^m(\hat{\mathbf{z}}_{s_m}^{\boldsymbol{\Phi}},s_m,\mathbf{y})-\boldsymbol{\Phi}(\tilde{\mathbf{z}}_{s_m}^{\boldsymbol{\Phi}},s_m, s_m,\mathbf{y})\\
                &=\hat{\mathbf{z}}_{s_m}^{\boldsymbol{\Phi}}-\tilde{\mathbf{z}}_{s_m}^{\boldsymbol{\Phi}}=\mathbf{z}_{s_m}-\mathbf{z}_{s_m}=0,    
            \end{aligned}
        \end{equation}
Therefore, 
\begin{equation}
    \begin{aligned}
        &||\boldsymbol{e}_n||\leq||\boldsymbol{e}_{s_m}||+L\sum_{t_i,t_{i-1}\in [s_m,s_{m+1}]} (t_i-t_{i-1}) O((\Delta t)^p)\\
        &= O((\Delta t)^p)\cdot(s_{m+1}-s_m).
    \end{aligned}
\end{equation}
The proof is completed.
\end{proof}
According to \cref{eq:sdslosswocfg}, one can ideally optimize a 3D model $\boldsymbol{\theta}$ such that $\mathbf{z}_{s_m}=\boldsymbol{G}_{\boldsymbol{\theta}}^m(\tilde{\mathbf{z}}_t^{\boldsymbol{\Phi}},t,\mathbf{y})$. In this case, we have $\boldsymbol{\Phi}(\tilde{\mathbf{z}}_{s}^{\boldsymbol{\Phi}},s, s_m,\mathbf{y})=\mathbf{z}_{s_m}^{data}$, where $\mathbf{z}_{s_m}^{data}=\alpha_{s_m}\mathbf{z}^{data}+\sigma_{s_m}\boldsymbol{\epsilon}^*$. Based on this, we have
\begin{equation}
    ||\boldsymbol{G}_{\boldsymbol{\theta}}^m(\tilde{\mathbf{z}}^{\boldsymbol{\Phi}}_{t},t,\mathbf{y})-\boldsymbol{\Phi}(\tilde{\mathbf{z}}_{t}^{\boldsymbol{\Phi}},t, s_m,\mathbf{y})||\!=\!||\mathbf{z}_{s_m}\!-\!\mathbf{z}^{data}_{s_m}||\!=\!||\mathbf{z}_0\!-\!\mathbf{z}^{data}||.
\end{equation}
Since we use a first-order ODE solver to implement $\boldsymbol{\Phi}$, we have
\begin{equation}
    \sup_{t,s\in[s_m,s_{m+1})}||\mathbf{z}_0 - \mathbf{z}^{data}||\\
          =\mathcal{O}(\Delta t)(s_{m+1}-s_m).
  \end{equation}
  The proof is completed.
% Having the supplementary compiled together with the main paper means that:
% \begin{itemize}
% \item The supplementary can back-reference sections of the main paper, for example, we can refer to \cref{sec:intro};
% \item The main paper can forward reference sub-sections within the supplementary explicitly (e.g. referring to a particular experiment); 
% \item When submitted to arXiv, the supplementary will already included at the end of the paper.
% \end{itemize}
% % 
% To split the supplementary pages from the main paper, you can use \href{https://support.apple.com/en-ca/guide/preview/prvw11793/mac#:~:text=Delete%20a%20page%20from%20a,or%20choose%20Edit%20%3E%20Delete).}{Preview (on macOS)}, \href{https://www.adobe.com/acrobat/how-to/delete-pages-from-pdf.html#:~:text=Choose%20%E2%80%9CTools%E2%80%9D%20%3E%20%E2%80%9COrganize,or%20pages%20from%20the%20file.}{Adobe Acrobat} (on all OSs), as well as \href{https://superuser.com/questions/517986/is-it-possible-to-delete-some-pages-of-a-pdf-document}{command line tools}.

\section{Visual Comparisons of State-of-the-arts}
\label{app:visual}
\begin{figure*}[!t]
    \centering
     \includegraphics[width=\linewidth]{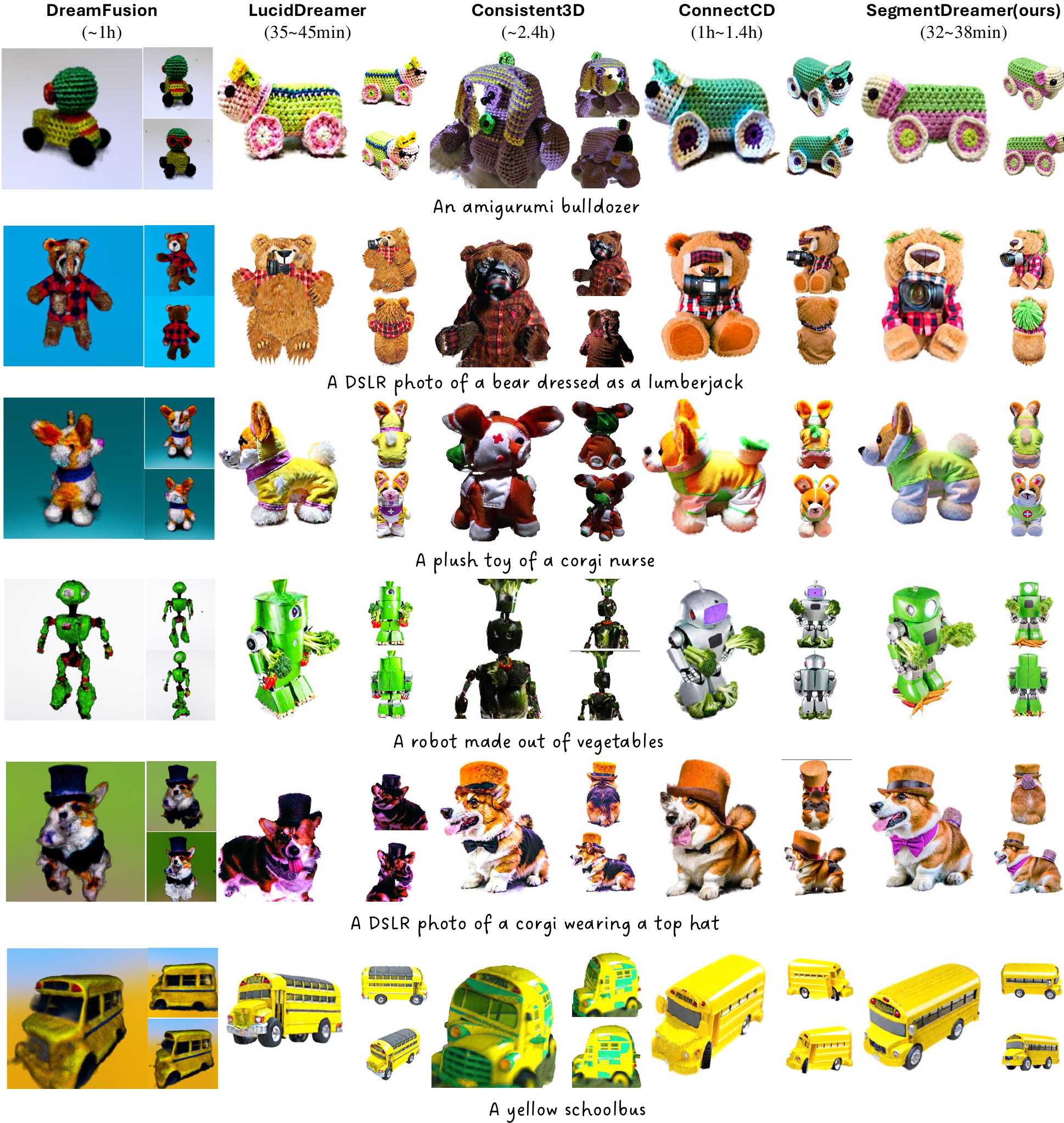}
     \caption{Additional qualitative comparisons with DreamFusion \cite{dreamfusion}, LucidDreamer \cite{luciddreamer}, Consistent3D \cite{consistent3d}, and ConnectCD \cite{connectingconsistency}. CFG scales are set to 100, 7.5, 20$\sim$40, 7.5, 7.5, respectively. Our approach yields results with high quality. Please zoom in for details.}
     \label{fig:app_compare}
  \end{figure*}

\begin{figure*}[!t]
  \centering
   %\includegraphics[width=8.6cm,height=4.2cm]{fig/framework.pdf}
   \includegraphics[width=\linewidth]{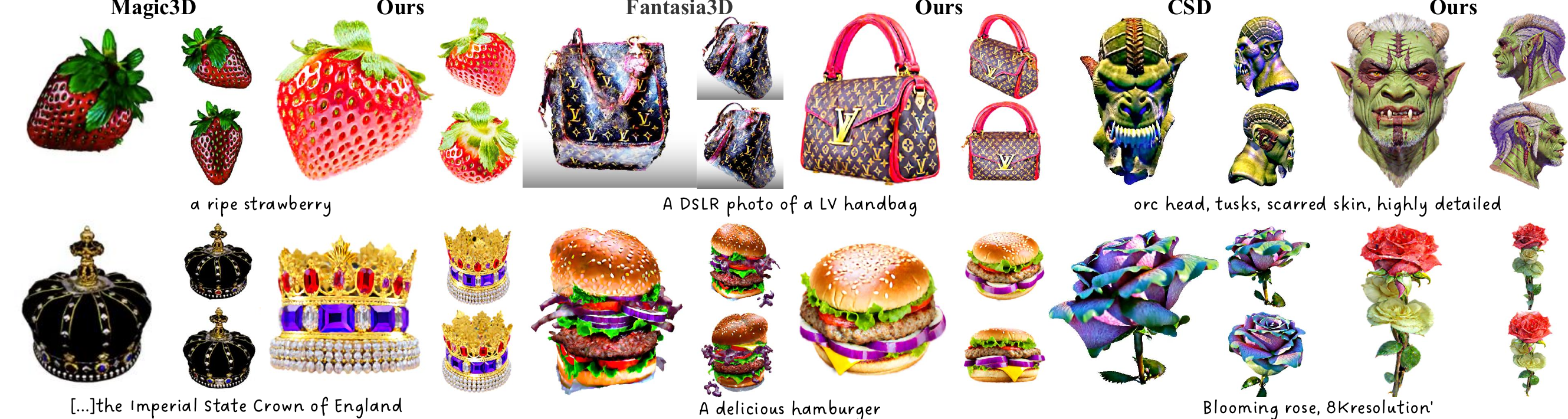}
   \caption{Additional qualitative comparisons with Magic3D \cite{dreamfusion}, Fantasia3D \cite{luciddreamer}, and CSD \cite{consistent3d}. Our approach yields results with high quality. Please zoom in for details.}
    \label{rebuttal:comp3}
    % \vspace{-2em}
\end{figure*}

We also present additional visual comparisons with state-of-the-art methods, as shown in \cref{fig:app_compare} and \cref{rebuttal:comp3}.
We provide additional qualitative comparisons against DreamFusion \cite{dreamfusion}, LucidDreamer \cite{luciddreamer}, Consistent3D \cite{consistent3d}, ConnectCD \cite{connectingconsistency}, Magic3D \cite{magic3d}, Fantasia3D \cite{fantasia3d}, and CSD \cite{csd}. As shown, our method clearly outperforms others visually.
% In Table \ref{tab:prompt}, we provide the detailed prompts used in the examples. 
% \section{Limitation and Potential Negative Impact}
% \label{app:limit}

% \noindent\textbf{Limitation.} SegmentDreamer is primarily designed for single-instance generation and performs suboptimally in multi-instance generation.

% \noindent\textbf{Potential Negative Impact.} SegmentDreamer focuses on content synthesis, which could potentially be misused to create misleading or deceptive content.
\begin{table*}[htbp]
  \centering
  \caption{40 prompts for evaluation.}
  \label{tab:prompts}
  \begin{tabular}{p{0.45\linewidth}p{0.45\linewidth}}  % 两列，各占45%宽度，适配文本换行
    \toprule
    \textbf{Column 1} & \textbf{Column 2} \\
    \midrule
    1. A cat with a mullet & 21. A blue motorcycle \\
    2. A pig wearing a backpack & 22. Michelangelo style statue of an astronaut \\
    3. A DSLR photo of an origami crane & 23. A DSLR photo of a chow chow puppy \\
    4. A photo of a mouse playing the tuba & 24. A DSLR photo of cats wearing eyeglasses \\
    5. An orange road bike & 25. A red panda \\
    6. A ripe strawberry & 26. A DSLR photo of an elephant skull \\
    7. A DSLR photo of the Imperial State Crown of England & 27. An amigurumi bulldozer \\
    8. A photo of a wizard raccoon casting a spell & 28. A typewriter \\
    9. A DSLR photo of a corgi wearing a top hat & 29. A red-eyed tree frog, low poly \\
    10. A rabbit, animated movie character, high-detail 3D model & 30. A DSLR photo of a chimpanzee wearing headphones \\
    11. A panda rowing a boat & 31. A robot made out of vegetables \\
    12. A highly detailed sand castle & 32. A DSLR photo of a red rotary telephone \\
    13. A DSLR photo of a chimpanzee dressed like Henry VIII king of England & 33. A DSLR photo of a blue lobster \\
    14. A photo of a skiing penguin wearing a puffy jacket, highly realistic DSLR photo & 34. A DSLR photo of a squirrel flying a biplane \\
    15. A blue poison-dart frog sitting on a water lily & 35. A DSLR photo of a baby dragon hatching out of a stone egg \\
    16. A DSLR photo of a bear dressed in medieval armor & 36. A DSLR photo of a bear dancing ballet \\
    17. A DSLR photo of a squirrel dressed like a clown & 37. A plate of delicious tacos \\
    18. A plush toy of a corgi nurse & 38. A DSLR photo of a car made out of cheese \\
    19. A humanoid robot playing the violin & 39. A yellow school bus \\
    20. A DSLR photo of a bear dressed as a lumberjack & 40. A DSLR photo of a shiny beetle \\
    \bottomrule
  \end{tabular}
\end{table*}
% \begin{table}
%     \centering
%     \begin{tabular}{cc}
%          & \\
%          & \\
%     \end{tabular}
%     \caption{Caption}
%     \label{tab:my_label}
% \end{table}
% {
%     \small
%     \bibliographystyle{ieeenat_fullname}
%     \bibliography{main}
% }
% \small
% \bibliographystyle{ieeenat_fullname}
% \bibliography{supp}

\end{document}